\documentclass{article}

\usepackage{ijcai22}

\usepackage[utf8]{inputenc} 
\usepackage[T1]{fontenc}    
\usepackage{hyperref}       
\usepackage{url}            
\usepackage{booktabs}       
\usepackage{amsfonts}       
\usepackage{nicefrac}       
\usepackage{microtype}      
\usepackage{xcolor}         
\usepackage{graphicx}
\usepackage{amsmath}
\usepackage{subfigure}
\usepackage[capitalize]{cleveref}
\usepackage{multirow}
\usepackage[normalem]{ulem}
\useunder{\uline}{\ul}{}

\title{Model-Contrastive Learning for Backdoor Defense}

\author{
 Zhihao Yue,
 Jun Xia,
 Zhiwei Ling,
 Ming Hu,
 Ting Wang,
 Xian Wei,
 Mingsong Chen$^*$\\
 $^1$Shanghai Key Lab of Trustworthy Computing, East China Normal University\\
$^*$Corresponding Author, \{mschen\}@sei.ecnu.edu.cn
}

\begin{document}

\maketitle

\begin{abstract}
  Due to the   popularity  
  of Artificial Intelligence (AI) techniques,  
  we are witnessing an increasing number  of 
  backdoor injection attacks that are designed to 
  maliciously threaten Deep Neural Networks
  (DNNs) causing  misclassification.
  Although there exist various defense 
  methods that can effectively erase backdoors from DNNs, they  
   greatly suffer from both  high
   Attack Success Rate (ASR) and  a  non-negligible loss in Benign Accuracy (BA).   
Inspired by the observation that 
   a  backdoored DNN tends to 
    form a  new cluster  in its  
    feature spaces for 
   poisoned data,  in this paper we propose a novel 
   two-stage backdoor defense method, named  MCLDef, based on Model-Contrastive Learning (MCL).
  In the first stage, our approach  
   performs   trigger inversion
   based on   trigger synthesis, where the resultant 
    trigger can be used  
   to generate poisoned data. 
In the second stage, 
under the guidance of  MCL and our defined positive and negative pairs, 
MCLDef can purify the backdoored
model by pulling the 
feature representations of poisoned data towards  those of 
their clean data counterparts.
Due to  the shrunken  cluster  of poisoned data, 
the backdoor   formed by end-to-end supervised learning 
is eliminated.
Comprehensive experimental results show  that, 
with  only 5\% of clean data,
MCLDef  significantly 
 outperforms  state-of-the-art 
 defense methods by up to 95.79\% reduction in ASR, while 
 in most cases the BA degradation can be controlled within less than 2\%. Our code is available at \url{https://github.com/WeCanShow/MCL}. 
 

   
\end{abstract}

\section{Introduction}


Along with   the prosperity of Artificial Intelligence (AI),
Deep Neural Networks (DNNs) play a more and more important role in the design of
 safety-critical systems (e.g., autonomous driving \cite{autonomous_driving},  medical diagnosis \cite{WangPLLBS17} and plant control \cite{ShiXXQ22}) for the purposes of intelligent sensing and control. However, due to 
 the steadily increasing  popularity,  DNNs are inevitably
 becoming the focus of various malicious adversaries \cite{song2020fda}. 
For example,  due to  biased third-party
 training data   or 
improper training processes,  a tiny adversarial 
input perturbation during the test phase can fool a target
DNN  to make  misclassification \cite{Carlini017}, resulting in 
disastrous consequences. Recently, 
backdoor attacks \cite{badnets} have emerged as a notable 
security threat to DNNs.
By poisoning partial
training data, adversaries 
can embed 
backdoors
in some DNN by tricking it
to establish the correlations between trigger patterns and 
target labels. In this way, 
 if some trigger pattern appears at test time, the DNN will make a preset prediction on purpose. 
In response to the 
situation of increasing popularity of backdoor attacks, various backdoor defense methods have been  developed, 
which can be mainly classified into two categories. 
The first one is the 
detection-based methods \cite{zheng2021topological,hayase2020spectre,NC}, 
which can effectively
figure out whether  training data are poisoned 
or  training processes are manipulated. 
However, these methods  focus 
on the detection rather than  elimination
of triggers within poisoned data or backdoors within DNNs. 
In this paper, we focus on the second one, i.e., 
erasing-based methods \cite{li2021neural,li2021anti,qiao2019defending} 
whose goal is to eliminate the 
backdoor triggers by purifying the
malicious effects of backdoored DNNs.

  




Although existing
erasing-based methods are promising in
fixing backdoored DNNs, most of them face  with the dilemma that 
the reduction of  Attack Success Rate (ASR) for DNNs 
may notably  sacrifice their 
  Benign Accuracy (BA) on clean data.
The reason of such low BA is mainly due to the 
  side effect of purification processes, 
  which messes up the  feature space of backdoored DNNs.
According to the observation in \cite{DBD}, 
due to   excessive learning capability of the end-to-end 
supervised learning paradigm, 
the features of backdoor triggers can be learned. As a result,  
poisoned data prefer to form a new cluster
in the feature space of a backdoored DNN
rather than  to locate
in the ones of 
benign samples.
However, when conducting some kind of 
 purification on the backdoored DNN, the  boundaries
of benign data clusters become blur, resulting in 
the low BA. 
Although  
the decoupling-based backdoor defense approach introduced in 
\cite{DBD} is the first to study
backdoors from the perspective of 
hidden feature space, it focuses on the construction of DNNs that are 
immune to backdoor attacks rather than eliminating backdoors from DNNs. 
When more and more AI applications adopt  pre-trained DNNs from untrusted  third
parties,  {\it how to 
fully take the advantage of  feature
distribution characteristics to 
enhance the capability of erasing
backdoor triggers without degrading BA  
is becoming a challenging direction in backdoor defense. 
}

 


Contrastive learning methods \cite{simclr,chen2020big,he2020momentum,chen2021exploring}   have been acknowledged as a promising way to learn better 
data feature representations.
By learning from both   positive pairs constructed from different transformations of 
same samples and negative pairs generated from different samples, 
contrastive learning can make the features within a positive pair  close to each other, 
and make
the two features within a negative pair  far away from each other.
Therefore, 
 contrastive learning is the best fit for  backdoor elimination from the perspective of feature distributions,
since it can  shrink the cluster of poisoned data as much as possible while 
making the feature representations close to
their benign counterparts.
Based on this understanding, 
in  this paper we propose a novel Model-Contrastive Learning (MCL)-based
backdoor defense method, named  MCLDef, which can make   full use of the 
distribution characteristics of both poisoned and benign samples 
in the feature space for the purpose of 
purifying  backdoored DNNs. 
The major contributions of this paper are as follows:
\begin{itemize}
\item Based on 
the distribution characteristics of feature representations,
we define both   positive pairs and negative pairs for data features
that  are specific for  MCL-based backdoor defense.

\item We propose a 
 two-stage  MCL framework for backdoor defense, 
which can 
effectively shrink or even damage the cluster of poisoned data in the 
feature space by pulling the feature representations of 
poisoned data towards to their benign counterparts.

\item 
We conduct  experiments against  
well-known backdoor attacks to show the superiority of MCLDef over 
state-of-the-art (SOTA)  backdoor defense methods in terms of ASR and BA.  

\end{itemize}

The rest of this paper is organized as follows. 
After the introduction to related work on backdoor inject attack, 
backdoor defense and contrastive learning in Section 2, 
Section 3  presents the detail of our  MCLDef approach. Section 4 
evaluates the effectiveness of our approach against four 
SOTA backdoor attacks using various well-known benchmarks. 
Finally, Section 5 concludes the paper.


\section{Related Work}

\textbf{Backdoor Attack.}
Generally, a backdoor attack refers to embedding
a backdoor
into
some DNN by poisoning part of its 
training data, where  triggers can be controlled  by 
adversaries to manipulate DNN outputs, causing incorrect or unexpected  behaviors.  
As the most common backdoor  attack paradigm, 
poisoned-label  attacks \cite{badnets,trojann}
stamp  backdoor triggers onto 
benign training
data and  replace 
their labels with attacker-specified ones. 
For example, Chen et al. \cite{physics_backdoor} introduced
the {\it blended attack} that 
 generates poisoned instances and backdoor instances
by blending  benign input data with a specific key pattern.
In practice, to avoid the early detection of backdoor attacks 
from observable training data, most attackers prefer to adopt 
 backdoor attacks with imperceptible changes, e.g., 
natural reflection \cite{liu2020reflection} and human imperceptible noise \cite{zhong2020backdoor}.
As  an alternative, clean-label attacks only modify partial training data
without  tempering their labels. 
For example, Turner et al. \cite{CL} proposed the {\it clean-label attack} 
that adopts
adversarial perturbations or generative models to tamper with 
 benign data from the target class and then performs trigger injection, thus 
establishing the connection between  triggers and their target labels.
In \cite{Sig}, Barni et al. 
implemented a clean-label  attack using sinusoidal bands as the trigger.
Note that, although 
clean-label  attacks are  more  stealthy than 
poisoned-label  attacks, they greatly  suffers from the problems of low 
ASR or even failures in creating backdoors \cite{li2020backdoor}.

\textbf{Backdoor Defense.}
The mainstream
backdoor defense methods can be mainly classified into 
two categories, i.e., the detection-based methods and 
the erasing-based methods. 
The former one can be used to: i) identify
backdoors  of DNNs  during the training \cite{zheng2021topological,dong2021black,xiang2022post}; 
or ii) check  whether the data are poisoned \cite{hayase2020spectre,zeng2021rethinking,tang2021demon}, thus 
the poisoning data can be safely filtered during the training. 
The latter category
consists of three directions: 
i) {\it model reconstruction-based defense}  that leverages
DNN finetuing \cite{li2021neural,liu2018fine,zeng2021adversarial} using  
a limited set of clean data 
to eradicate
backdoor triggers; 
ii) {\it poison suppression-based defense}  that mitigates 
the impacts of poisoned
samples during the training process  \cite{li2021anti,DBD,borgnia2021strong}; 
and iii) {\it trigger synthesis-based defense} that 
uses  trigger synthesis  to construct poisoned samples
for the purpose of backdoor trigger elimination  \cite{NC,qiao2019defending}. 


\textbf{Contrastive Learning.} Contrastive learning has been widely investigated
in self-supervised learning to  achieve 
desired  feature representations from unlabeled samples  
\cite{simclr,chen2020big,he2020momentum,chen2021exploring,jing2020self}.
Its basic idea  is to 
reduce the  distance of
feature representations of two samples within a positive pair 
 as much as possible, and
enlarge the  distance of feature representations of different samples 
within negative pairs.  
Besides the contrastive learning on samples,  Model-Contrastive Learning  (MCL)
has been studied
to correct the bias of  local models for federated learning \cite{moon}.

Our approach is inspired by the observation in \cite{DBD}, which has been used to 
construct DNNs that are immune to backdoor attacks by decoupling  training processes. However, this method cannot be directly used to purify 
backdoored DNNs. To the best of our knowledge, MCLDef is the first attempt that leverages the merits of MCL to eliminate backdoors in DNNs. By correcting  the feature representations of  poisoned samples, MCLDef
can not only mitigate the malicious impacts of backdoor attacks, but also 
maintain an expected  classification performance with high BA.

\begin{figure*}[t]
  \centering
  \hspace{-0.25in} \subfigure[BadNets\label{motivation:Badnets}]{\includegraphics[width=1.23in]{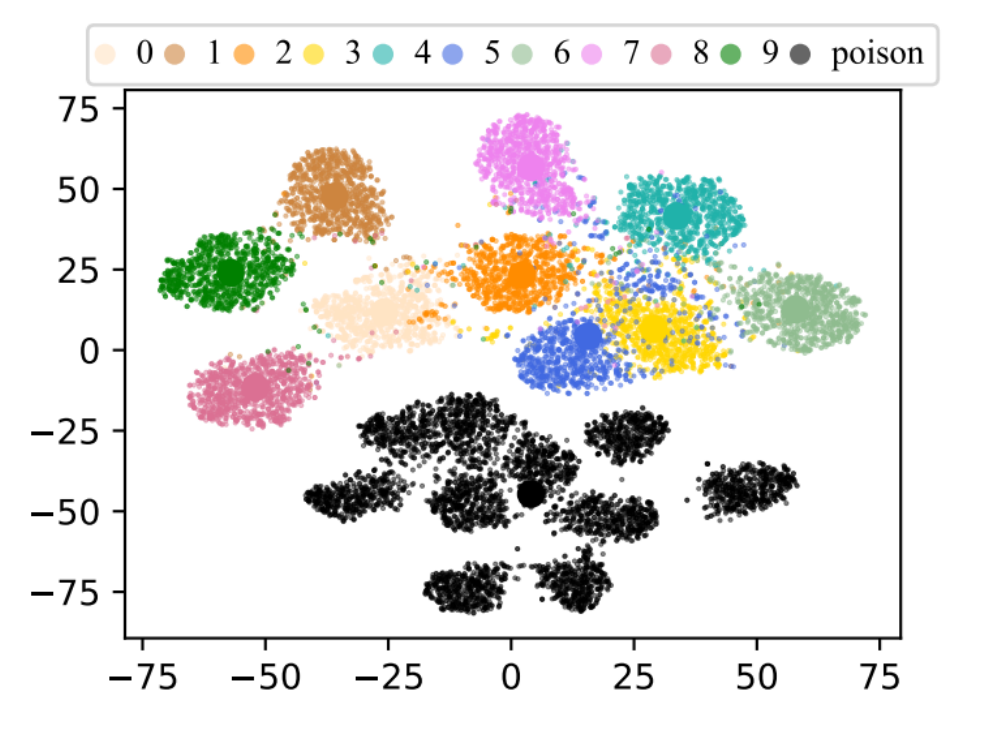}}
  \hspace{-0.18in}
  \subfigure[Trojan\label{motivation:Trojan}]{\includegraphics[width=1.23in]{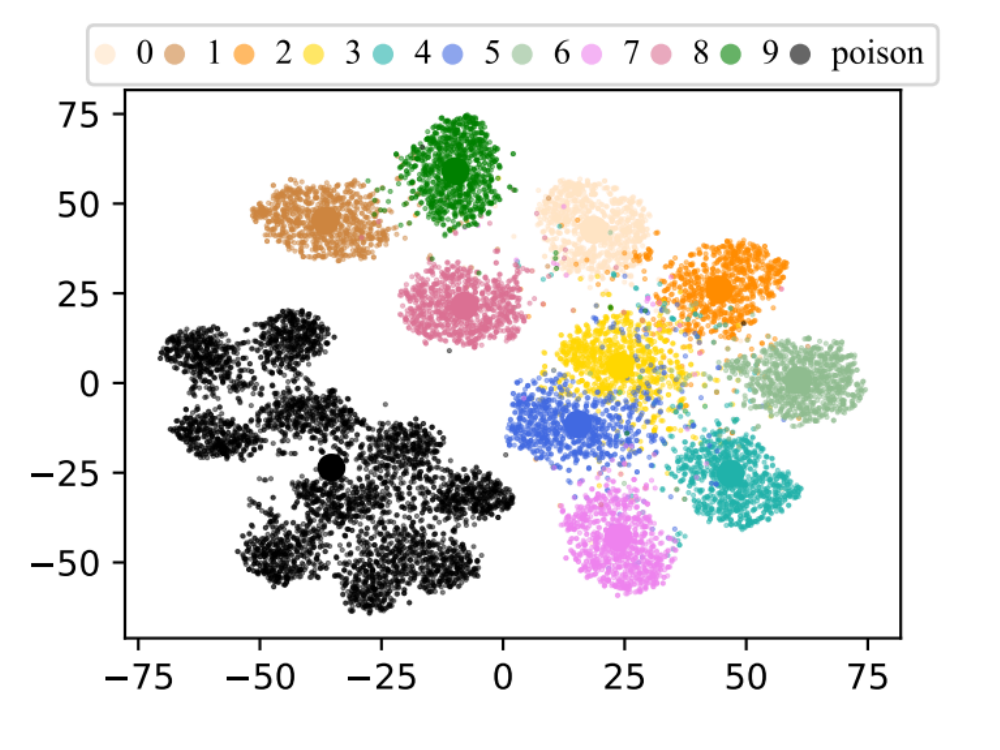}} 
    \hspace{-0.18in}
   \subfigure[Blend\label{motivation:Blend}]{\includegraphics[width=1.23in]{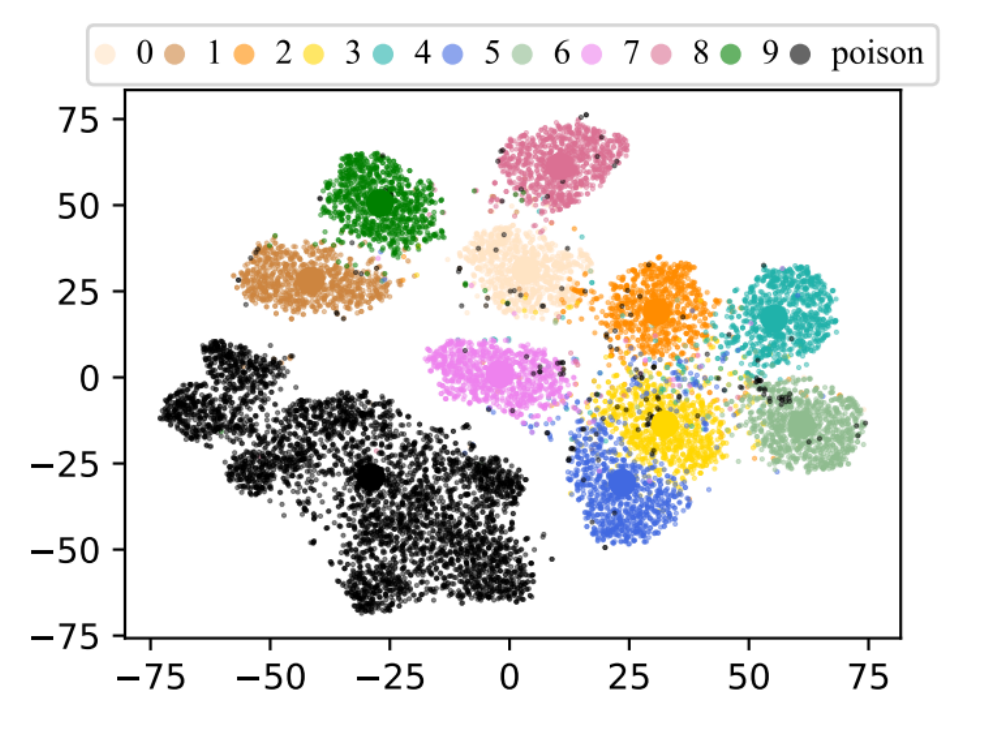}} 
    \hspace{-0.18in}
   \subfigure[CL\label{motivation:CL}]{\includegraphics[width=1.23in]{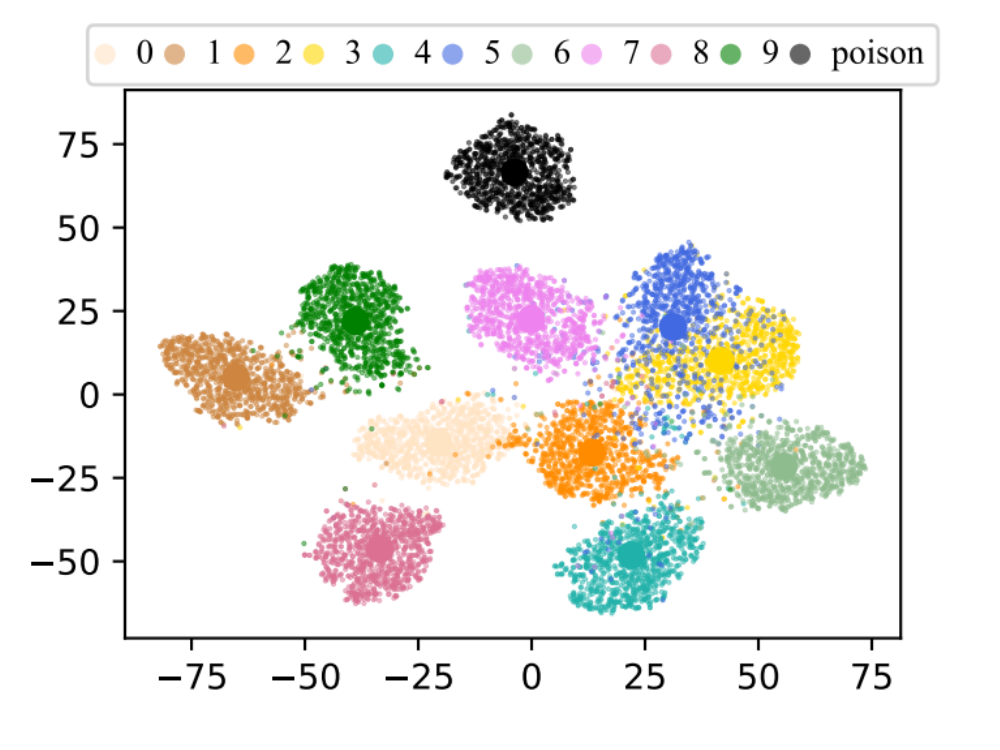}} 
     \hspace{-0.18in}
   \subfigure[Sig\label{motivation:Sig}]{\includegraphics[width=1.23in]{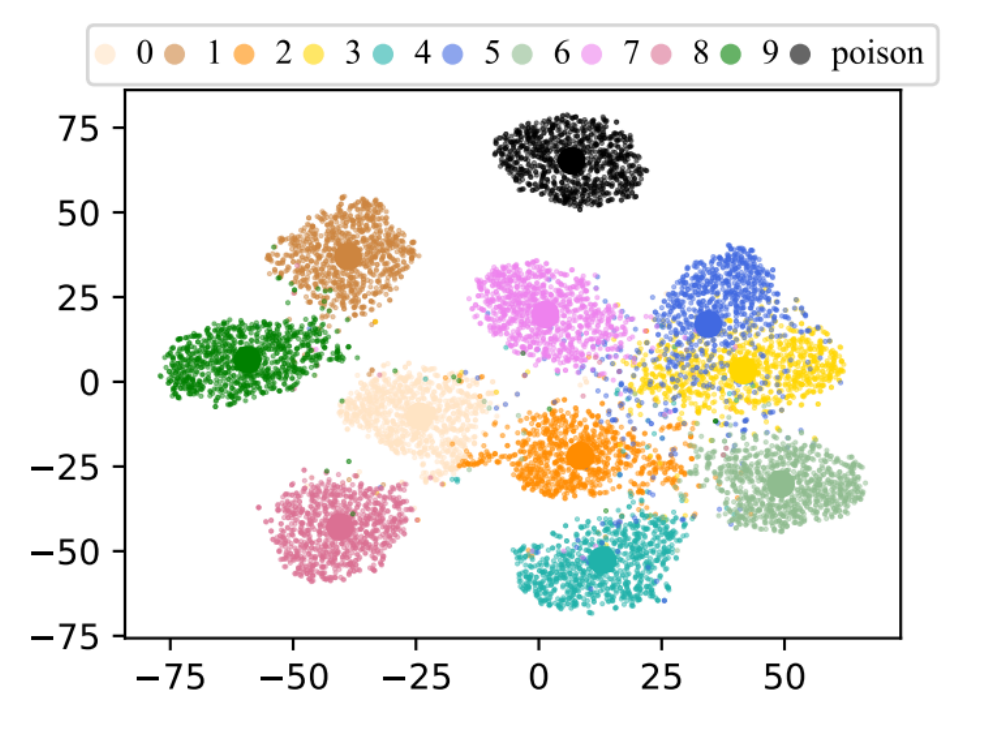}} \\
  \caption{The t-SNE  of  feature vectors in the hidden feature space under different attacks.} \label{figure_motivation}
\end{figure*}

\section{Our MCL-based  Backdoor Defense Method}

In this section, we first show the motivation of our approach from the perspective of feature distributions, and then introduce our two-stage MCL-based backdoor defense method in detail.


\textbf{Defense Settings.}
Similar to other classical backdoor defense methods, in this paper we assume that 
a backdoored DNN is trained by an unreliable third-party dataset,
where attackers can  arbitrarily modify the training dataset but cannot control the 
training process. Meanwhile, 
we assume that partial (e.g., 5\%)
clean training data is available to finetune the backdoored model. 
The goal of MCLDef is to use such
clean data to purify the poisoned DNN by erasing the 
injected backdoor without degrading the classification performance on benign data.

\subsection{Motivation from Hidden Feature Space}
\label{motivation}

To understand the distribution characteristics of feature representations in the hidden feature space, 
we  evaluate the effects of  five backdoor injection attacks, i.e., 
BadNets \cite{badnets}, Trojan \cite{trojann}, Blend \cite{physics_backdoor}, CL \cite{CL}, and Sig \cite{Sig}, respectively, on Cifar-10 dataset \cite{krizhevsky2009learning}. 
Here, we obtain feature vectors of the test result from the feature extractor (the DNN without the last 
classifier layer) and use t-SNE \cite{van2008visualizing} for visualization. Please refer to 
Appendix~\ref{appendix_reproduce_backdoor}  for more details about the  experimental settings. 
Figure~\ref{figure_motivation} shows the distributions of feature representations of both 
benign and poisoned samples under the five attacks respectively. For different kinds of 
input training
samples, we marked 
their feature representations in different colors (i.e., color 0 to 9). Note that in each subfigure the feature representations of 
poisoned data   are marked
in black color. 
From this figure, we can clearly find that the feature representations of 
benign samples from the same category of Cifar-10 form an individual cluster, while all the poisoned samples form a new cluster (in black color) in the feature space. According to the observation in
\cite{DBD}, this new cluster implicitly
reflects the features of the injected trigger itself.
Note that the attacks
in Figure \labelcref{motivation:Badnets,motivation:Trojan,motivation:Blend}
are all  poisoned-label backdoor attacks. From these three subfigures, we 
can observe that the black cluster can be 
further divided into 10 subclusters, which correspond to the 10 clusters of their benign counterparts. 
This phenomenon again demonstrates  that the sample feature  is also learnt by the DNN. 
Unlike the three attacks, the ones used 
in Figure \labelcref{motivation:CL,motivation:Sig} are all clean-label  attacks, 
where we can only  find one single  dense
black cluster for the poisoned samples. This is because 
clean-label  methods  only attack the samples with 
a specific target-label (i.e., the label that corresponds to blue color  in this example).


Based on the above observations, we can find that the cluster of poisoned samples 
indicates the trigger pattern for backdoor attack from the perspective of feature space. 
In other words, if we can shrink or damage such a 
cluster of poisoned samples, we can eliminate the backdoor triggers as well. 
However, this may cause the problem of low BA, since the new features of poisoned 
samples will be scattered improperly to the incorrect places of the feature space.  
To avoid  BA degradation, it is required that the new features of poisoned samples after purification
can located close to their benign counterparts.

\subsection{Overview of  Our Approach}
\label{overview}

In this paper, we resort to MCL for the backdoor defense, which 
naturally matches the observations in Section \ref{motivation}. The basic idea of our approach is to 
define a benign sample and its poisoned counterpart as a positive pair and define 
two different benign samples as a negative pair. Under the guidance of 
MCL, the cluster of poisoned samples will be damaged, and the new features 
of  poisoned samples  will be pulled towards to the ones of  their benign counterparts. 
In this way, the backdoor triggers will be eliminated with negligible BA loss.

\begin{figure*}[ht]
\centering
\includegraphics[width=4.5in]{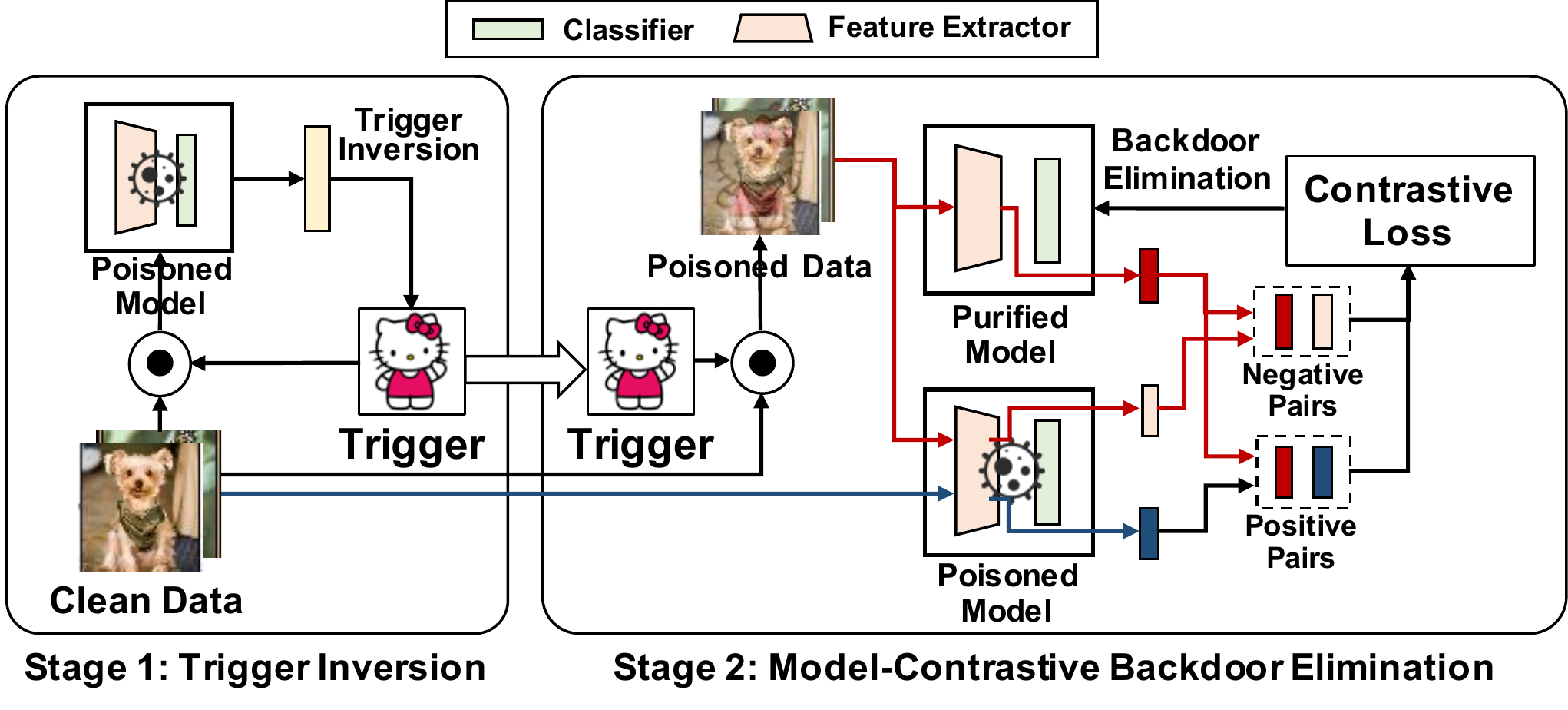}
\caption{Overview of our MCLDef defense method} \label{figure_overflow}
\end{figure*}

To enable MCL for backdoor defense, we need to figure out two questions first. 
The first one is ``how can we figure out the clusters for poisoned samples'', and the second one is 
``how can we accommodate the MCL to enable efficient backdoor defense without 
degrading BA''. To address these two questions, in this paper we propose a novel two-stage MCL-based 
backdoor defense method called MCLDef as shown in Figure \ref{figure_overflow}. The first stage
performs  trigger inversion based on an  existing
trigger synthesis method presented in \cite{NC} that converts the 
trigger inversion into an optimization problem. Based on the  poisoned samples generated 
by the obtained inverted trigger in the first stage, the second stage utilizes the model-based
constrastive learning to purify the backdoored DNNs based on our defined 
positive and negative pairs. 
Note that in our approach, we treat a DNN as 
a combination  of two disjoint parts including a feature extractor and a classifier (i.e, fully connected
layers), whereas the  positive and negative pairs are constructed based on the outputs of 
feature extractors of both the poisoned model and purified model. 
Based on the calculated contrastive loss, the backdoor elimination process iterates until
the convergence of purification.  The following subsections will detail the key components and 
steps of our approach.

\subsection{Trigger Inversion (Stage 1)}

As aforementioned, our approach 
 purifies backdoored DNNs based on both the positive and negative pairs of feature representations. 
Since our approach only uses a small set of clean samples for the purpose of finetuning, 
we need to figure out 
their poisoned counterparts. However, for a given backdoored DNN, it is hard to know which kind of 
backdoor attack is applied on the model. To address this problem, 
the first stage of our approach resorts to trigger inversion, which can effectively to figure out a trigger to derive   
 poisoned 
samples with high  quality. 
Here, we adopt the same trigger inversion implementation as the 
 well-known Neural Cleanse (NC) method \cite{NC}. 

Assume that  $\mathcal{D}=\left\{(\boldsymbol{x_i},y_i)\right\}_{i=1}^N$ is a clean dataset, where $\boldsymbol{x_i}\in\{0,1,...,255\}^{C \times W \times H}$.
We  generate the corresponding poisoned dataset $\mathcal{\hat{D}}=\left\{(\boldsymbol{\hat{x_i}},\hat{y_i})\right\}_{i=1}^N$ by using the following equation:
\begin{equation}    \label{eq_poi}
    \hat{\boldsymbol{x}}=(1-\boldsymbol{m})\boldsymbol{\bigodot{x}} + \boldsymbol{m}\bigodot{\boldsymbol{\Delta}},
\end{equation}
where $\boldsymbol{\Delta}$ denotes
the trigger pattern, $\boldsymbol{m}\in\left\{0,1\right\}^{W \times H}$ is a binary mask used to determine    the trigger position, and $\bigodot$ indicates
the element-wise  product. 
Assuming that the target label be $y_t$, our approach  optimizes
the trigger pattern $\boldsymbol{\Delta}$ and the mask $\boldsymbol{m}$ as follows:
\begin{equation}    \label{eq_nc}
\begin{split}
     \boldsymbol{\Delta^{*}},\boldsymbol{m^{*}}={\underset {\boldsymbol{\Delta},\boldsymbol{m}} { \operatorname {arg\,min} } \, \sum_{i=1}^{N}{\mathcal{L}(f(\boldsymbol{\hat{x_i}};\boldsymbol{\Bar{\omega}}),y_t)+\lambda|\boldsymbol{m}|} },
\end{split}
\end{equation}
where $\boldsymbol{\Bar{\omega}}$ is the poisoned model parameters. $f(\cdot)$ is the DNN output,  $\mathcal{L}(\cdot)$ is the cross-entropy loss function, and  $\lambda$ is a hyperparameter that controls the weight of the regularization. Here, 
we assume that the target label of  backdoor attacks has been  
detected by some method, e.g.,  the ones proposed in \cite{zheng2021topological,NC,dong2021black}.
Based on Equation \ref{eq_nc}, we can obtain an optimized 
 trigger pattern $\boldsymbol{\Delta^{*}}$ and its mask $\boldsymbol{m^{*}}$.

\subsection{Construction of Positive and Negative Pairs}    \label{construct_pair}

The positive pairs and negative pairs play an important role in 
MCL to determine the correlations between 
feature representations in feature space. By definition, after the contrastive learning,  the two
elements of a positive pair will become closer to each other , while the distance of 
two elements within a negative pair will become longer. 
Based on these facts, to accommodate MCL
to  the backdoor defense, we define the positive pairs by coupling 
the feature representation of a benign sample in the poisoned model and 
the feature representation of its poisoned counterpart
from the purified model together.
Since the poisoned model behaves normally on benign samples, in this way 
wen can  pull the feature representations of poisoned samples toward those of their clean sample  counterparts in the feature space of purified model, thus the BA can be guaranteed. 
Meanwhile, since we want to shrink or even totally damage the cluster of poisoned samples in the 
purified model, we define a negative pair by coupling the 
feature representations of a poisoned 
sample in both the poisoned model and purified model together. 
In this way, the purification can 
scatter the clustered 
 feature representations of 
poisoned samples within  feature space of  the purified model, thus damaging the 
trigger pattern of the purified model.

Formally, for a given benign sample $(\boldsymbol{x}, y )\in \mathcal{D}$, we can get the corresponding poisoned sample $(\boldsymbol{\hat{x}},\hat{y})\in \mathcal{\hat{D}}$ by Equation \ref{eq_poi}.
Let $h(\cdot)$ denote the feature vector output by the feature extractor (represented by the pink trapezoid in  Figure \ref{figure_overflow}).
Let $\boldsymbol{\omega}$ and $\boldsymbol{\Bar{\omega}}$ denote the 
parameters  of purified DNN and poisoned DNN, respectively, where  
$\boldsymbol{\omega}=\boldsymbol{\Bar{\omega}}$ at the beginning of 
the second stage. Assming that  $\boldsymbol{z}=h(\boldsymbol{\hat{x}};\boldsymbol{\omega})$, 
$\boldsymbol{z^{poi}}=h(\boldsymbol{\hat{ x}};\boldsymbol{\Bar{\omega}})$, 
and $\boldsymbol{z^{cle}}=h(\boldsymbol{x};\boldsymbol{\Bar{\omega}})$, 
we can construct a positive pair using $(\boldsymbol{z},\boldsymbol{z^{cle}})$  and a negative 
pair using $(\boldsymbol{z},\boldsymbol{z^{poi}})$.

\subsection{MCL-based Backdoor Elimination (Stage 2)}

As aforementioned, we can get the trigger pattern and the mask, which can be used to generate the poisoned data. According to the definition of positive and negative pairs in Section \ref{construct_pair}, we can get the positive pairs and negative pairs. However, the final goal is to eliminate the backdoor trigger and reduce the loss of the BA. Therefore, we need to accommodate the MCL to enable efficient backdoor defense without degrading BA. The second stage of our approach resorts to MCL, which can reduce the distance of feature representations of a positive pair as much as possible, while enlarging the distance of feature representations of a negative pair.

Figure \ref{figure_overflow} present the detailed structure of stage 2
in its right part. Once stage 1 finishes, we can obtain an optimized 
trigger pattern $\boldsymbol{\Delta^{*}}$ and its mask $\boldsymbol{m^{*}}$ from 
Equation \ref{eq_nc}. Assume that $\mathcal{D}$ is a clean dataset  with N clean samples and the derived trigger (i.e., the combination of $\boldsymbol{\Delta^{*}}$ and  $\boldsymbol{m^{*}}$). Based on these two things, we can generate a set of N poisoned samples using Equation \ref{eq_poi}, denoted as $\mathcal{\hat{D}}$.
Let $(\boldsymbol{x_i}, y_i)$ be a clean sample and 
$(\boldsymbol{\hat{x_i}}, \hat{y_i})$ be its corresponding poisoned sample. 
According to the definition of positive and negative pairs in Section \ref{construct_pair}, we can generate the corresponding positive pair $(\boldsymbol{z_i},\boldsymbol{z_i^{cle}})$  and negative pair $(\boldsymbol{z_i},\boldsymbol{z_i^{poi}})$ for   $\boldsymbol{x_i}$ and
$\boldsymbol{\hat{x_i}}$. 
Based on  the  NT-Xent loss function \cite{NT_Xent} 
that is commonly used as a template in contrastive learning and our proposed positive and negative pairs, we define our  model-contrastive loss for MCLDef  as follows:
\begin{scriptsize}
\begin{equation}    \label{contrastive_loss}
    \mathcal{L}_{contrastive}=\sum_{i=1}^{N}{-log\frac{exp(sim(\boldsymbol{z_i},\boldsymbol{z_i^{cle}})/\tau)}{exp(sim(\boldsymbol{z_i},\boldsymbol{z_i^{cle}})/\tau)+exp(sim(\boldsymbol{z_i},\boldsymbol{z_i^{poi}})/\tau)}},
\end{equation}
\end{scriptsize}
where $\tau$ is a temperature parameter that  controls the smoothness of  soft labels, and  $sim(\cdot,\cdot)$ represents the similarity of two feature vectors based on the cosine similarity function. In  stage 2, we optimize $\boldsymbol{\omega}$ via MCL  as follows:
\begin{equation}
    \boldsymbol{\omega^{*}}={\underset {\boldsymbol{\omega}} { \operatorname {arg\,min} } \, \mathcal{L}_{contrastive} }.
\end{equation}


\section{Experiments}       
\label{exps}

To show the effectiveness of our approach, 
we implemented  MCLDef  using  Pytorch, and 
 compared the performance of  MCLDef with
 four start-of-the-art (SOTA) backdoor elimination methods. 
 All the experimental results were obtained from a 
a workstation with 3.6GHz Intel i9 CPU, 32GB memory, NVIDIA GeForce RTX3090 GPU, and Ubuntu operating system. 
We designed comprehensive experiments to answer the following three research questions:

\textbf{RQ1 (Superiority over SOTA)}: What are the advantages of MCLDef compared with SOTA methods? 

\textbf{RQ2 (Effectiveness of MCL)}: Can MCL   improve the defense performance of 
 purified DNNs?

 \textbf{RQ3 (Applicability of MCLDef)}: What are the factors 
  (e.g., clean data rates, trigger inversion quality)
that greatly impact the performance of MCLDef?

\subsection{Experimental Settings}

\textbf{Datasets and DNNs.}
We evaluated all the  defense methods on three classical benchmark datasets, i.e.,  Cifar-10 \cite{krizhevsky2009learning}, GTSRB \cite{stallkamp2012man}, and an ImageNet subset (with first 20 categories) \cite{deng2009ImageNet}. We used WideResNet(WRN-16-1) \cite{zagoruyko2016wide} and ResNet18 \cite{he2016deep} as the DNNs for the experiments. The statistics of datasets and DNNs adopted in the experiments are presented in Table~\ref{dataset} (see Appendix~\ref{appendix_dataset}).

\textbf{Attack Configurations.}
We evaluated the performance of five attacks including four SOTA methods, 
which can be classified into two categories: i)  
poisoned-label  attacks including BadNets \cite{badnets}, Trojan \cite{trojann}, and Blend \cite{physics_backdoor}, where the poisoning  data rate is 10\% and 
the target label is $y_t=5$; and ii) 
clean-label  attacks including CL \cite{CL} and Sig \cite{Sig}, where 
the poisoning  data rate is 8\% and 
the target label is $y_t=5$.
We followed the settings suggested by \cite{li2021neural} to configure these attacks. 
Please see Appendix \ref{appendix_attack} for more details.


\textbf{Defense Configurations.}
We compared  MCLDef  with four SOTA backdoor defense methods, i.e.,  Fine-Tuning (FT) \cite{trojann}, Neural Attention Distillation (NAD) \cite{li2021neural}, Implicit Backdoor Adversarial Unlearning (I-BAU) \cite{zeng2021adversarial}, and 
Neural Cleanse (NC) with unlearning strategy \cite{NC}. Similar to the work in 
\cite{li2021neural}, we assumed that all
the methods could access  5\% clean training data for the finetuning. 
For  MCLDef, we adopted the same settings as the ones used by 
NC \cite{NC} for the implementation of trigger inversion.
Note that the initial purified model and poisoned model in model-contrastive 
backdoor elimination are the same. For the purified model, we set its  
 SGD optimizer with a momentum of 0.9, a weight decay of $1\times10^{-4}$, and an initial learning rate of 0.1 on both Cifar-10 and the ImageNet subset (0.01 on GTSRB). We used a batch size of 64 and finetuned the model for 20 epochs. 
 For the contrastive loss, we  set $\tau=0.5$. For fair comparison, 
  the settings of other four defense methods are consistent with  MCLDef. 
  Please see  Appendix~\ref{appendix_defense} for more 
   implementation details.


\textbf{Evaluation Metrics.}
Based on the clean test dataset, we used two metrics to 
evaluate the performance of  all the  defense methods: i)
Attack Success Rate (ASR) indicating  the 
ratio of succeed attacks over all the attacks on 
clean test samples without target labels; and ii) 
Benign Accuracy (BA) denoting the ratio of correctly 
classified samples over all the clean samples.

\begin{table*}[t]
\caption{Performance comparison between MCLDef and four SOTA defense methods}
\centering
\setlength{\tabcolsep}{0.7mm}{
\begin{tabular}{c|c|cc|cc|cc|cc|cc|cc}
\hline
\multirow{2}{*}{Dataset}  & \multirow{2}{*}{Attack} & \multicolumn{2}{c|}{Before} & \multicolumn{2}{c|}{FT} & \multicolumn{2}{c|}{NAD} & \multicolumn{2}{c|}{I-BAU} & \multicolumn{2}{c|}{NC}     & \multicolumn{2}{c}{MCLDef}         \\ \cline{3-14} 
                          &                         & ASR          & BA          & ASR     & BA           & ASR        & BA            & ASR        & BA        & ASR           & BA         & ASR           & BA            \\ \hline 
                          \hline
\multirow{5}{*}{Cifar-10} & BadNets                 & 99.64        & 83.01        & 12.78   & 80.64         & 5.76       & {\ul 81.71}          & {\ul 2.82}    & 79.54 & 3.58    & 81.37 & \textbf{2.22} & \textbf{82.19} \\
                          & Trojan                 & 99.38        & 82.84        & 17.13    & 80.93         & 5.14       & {\ul 81.20}     & 5.36    & 80.05  & {\ul 3.86}    & 79.61       & \textbf{2.62} & \textbf{82.04} \\
                          & Blend                   & 98.98        & 84.20        & 2.51    & 80.12          & 3.84 &  80.79    & {\ul 2.31}    & {\ul 81.54} & 3.01          & 80.94       & \textbf{2.12} & \textbf{83.98} \\
                          & CL                      & 95.47        & 86.44        & 6.7    & 80.34         & 6.99       &  81.39   & 13.62    & {\ul 82.15} & {\ul 5.71}    & 77.31        & \textbf{0.24} & \textbf{83.97} \\
                          & Sig                      & 99.03        & 84.64        & 68.17   & {\ul 80.98}   & {\ul 6.81} & \textbf{81.79} & 72.51    & 80.51 & 17.33         & 73.34       & \textbf{3.56} & 73.81          \\ \hline\hline
\multirow{3}{*}{GTSRB}    & BadNets                 & 100          & 97.86        & 13.25   & 92.04         & 4.85       & {\ul 92.89}    &  1.79    & 91.55 & \textbf{0.68} & 92.31       & {\ul 0.92}    & \textbf{96.32} \\
                          & Trojan                 & 99.38        & 98.08        & 20.99   & 92.13         & 8.54       & 91.94          & 23.73    & 90.12 & \textbf{0.54} & {\ul 92.37} & {\ul 2.77}    & \textbf{96.41} \\
                          & Blend                   & 99.04        & 97.79        & 10.63   & {\ul 93.13}   & 9.62       & 93.01          & 28.62    & 90.58 & {\ul 1.08}    & 91.39       & \textbf{0.32} & \textbf{96.22} \\ \hline\hline
\multirow{3}{*}{ImageNet}    & BadNets                 & 100          & 79.16        & 2.21   & 71.26         & 4.84       & {\ul 72.74}    &  4.1    & 41.37 & \textbf{0.84} & 72.42       & {\ul 1.79}    & \textbf{79.89} \\
                          & Trojan                 & 98.21        & 82.53        & 2.74   & 70.84         & 1.05       & 74.21          & 3.16    & 35.16 & {\ul 1.26} & {\ul 74.32} & \textbf{0.84}    & \textbf{83.37} \\
                          & Blend                   & 99.26        & 84.32        & 18.11   & {\ul 75.37}   & 14.32       & 74.32          & \textbf{1.36}    & 29.58 &  5.89    & 68.84       & {\ul 3.58} & \textbf{81.89} \\ \hline                    
                                   
\end{tabular}}   \label{comparison}
\end{table*}

\begin{figure*}[t]
\vspace{-0.15in}
  \centering
  \subfigure[BadNets against NAD\label{exp_motivation:NAD_badnet}]{\includegraphics[width=1.8in]{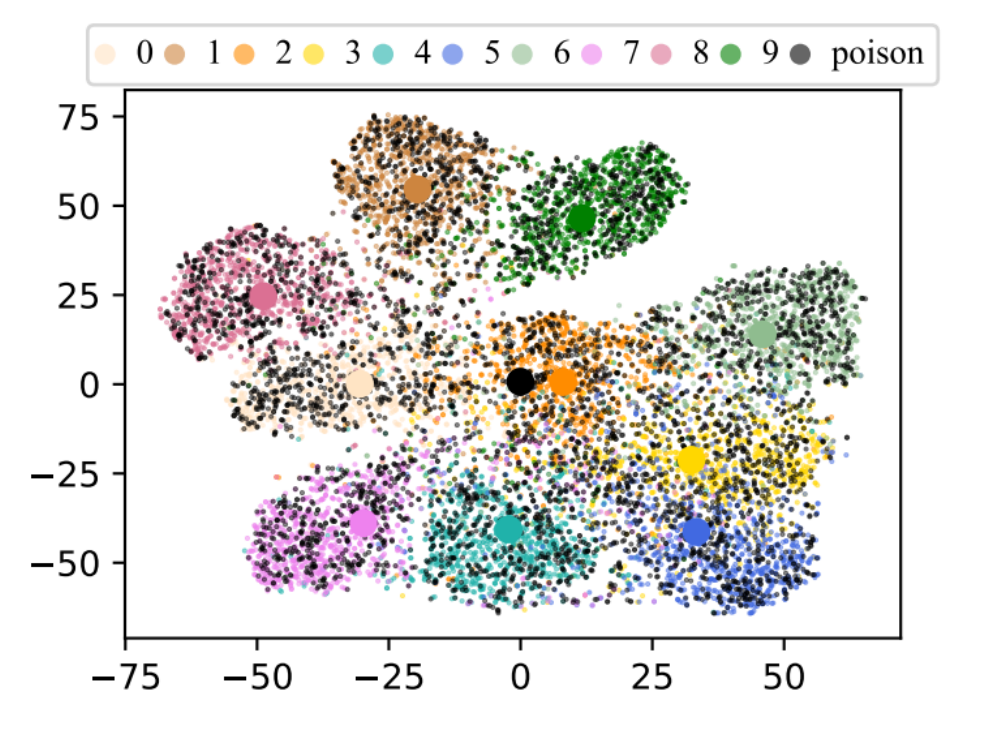}}  
  \subfigure[BadNets after NC\label{exp_motivation:NC_badnet}]{\includegraphics[width=1.8in]{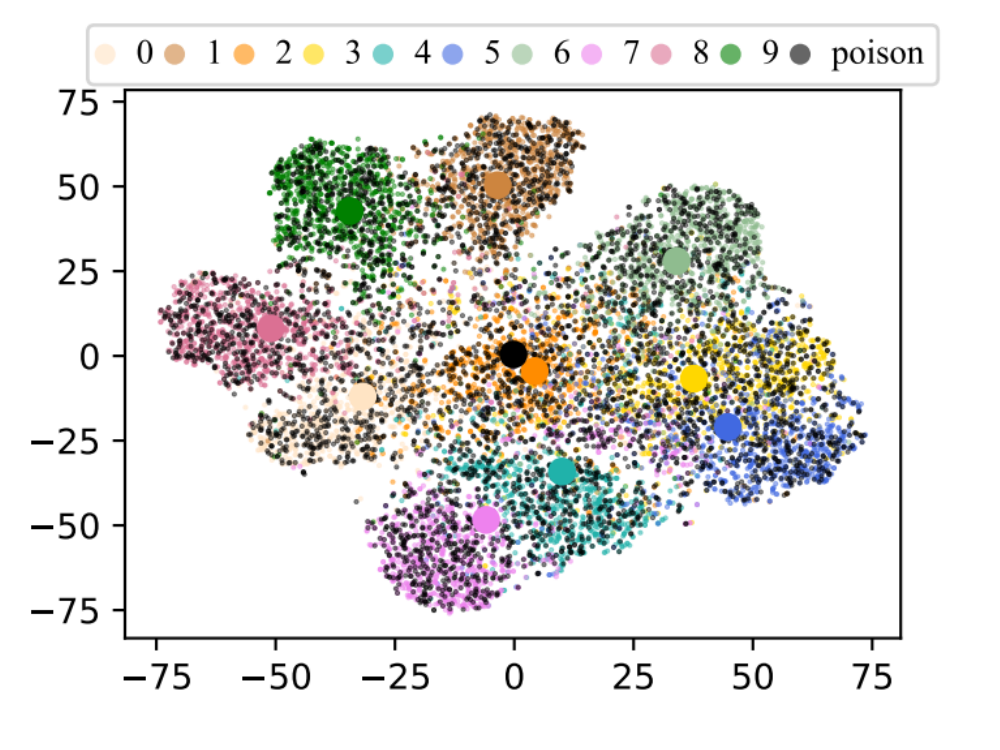}} 
   \subfigure[BadNets after MCLDef\label{exp_motivation:MCL_badnet}]{\includegraphics[width=1.8in]{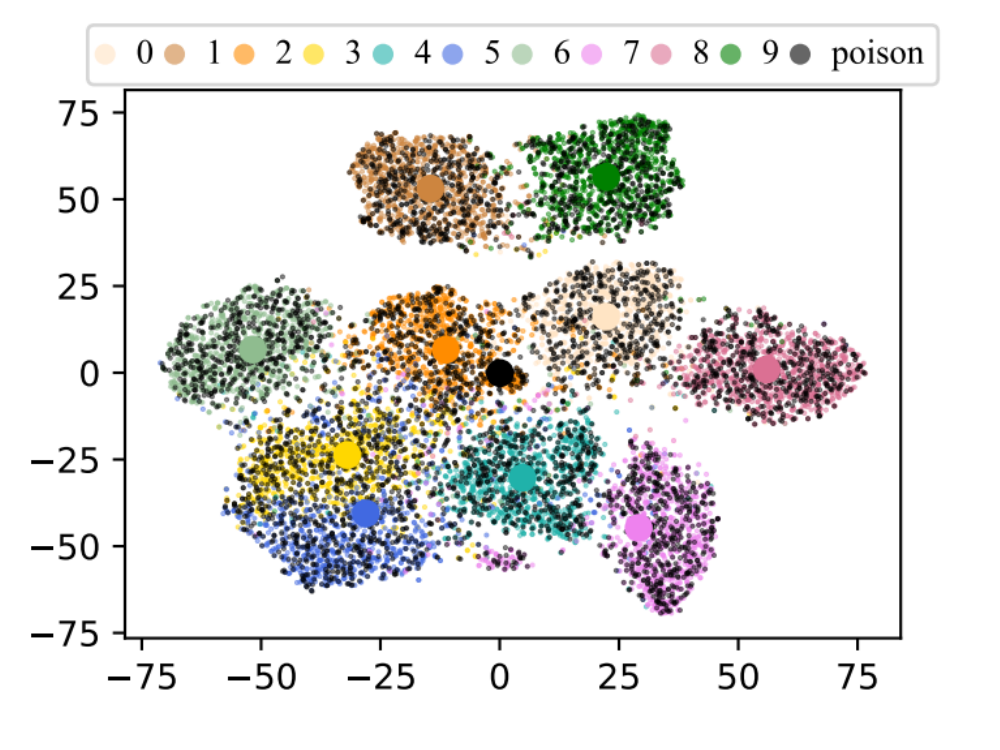}} \\
   \vspace{-0.05in}
    \subfigure[CL after NAD\label{exp_motivation:NAD_cl}]{\includegraphics[width=1.8in]{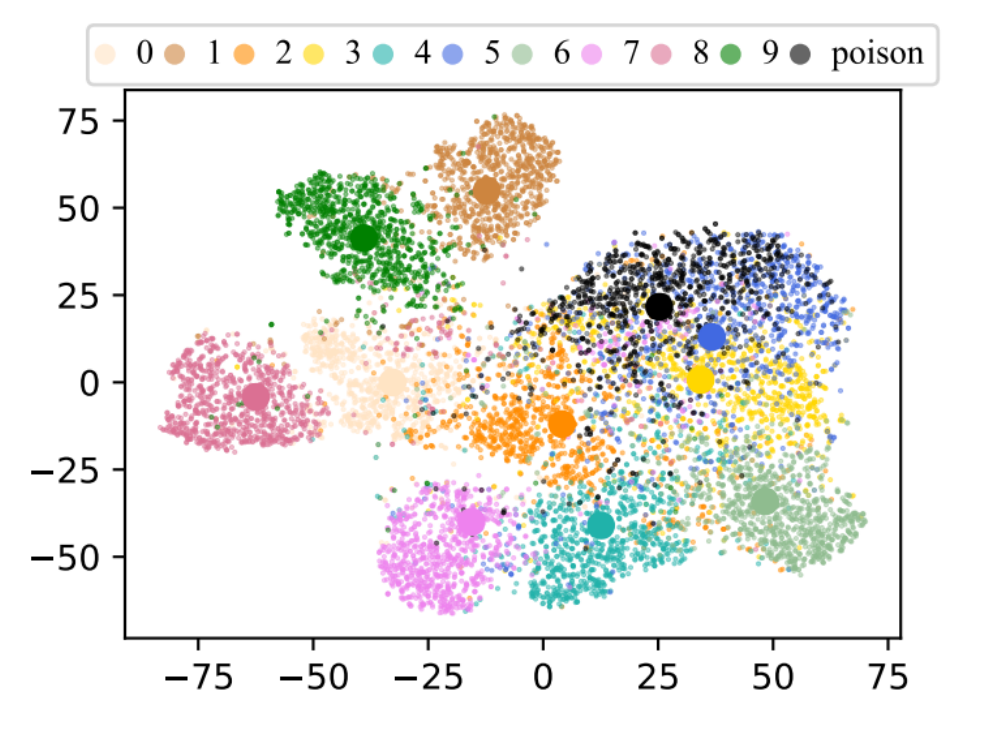}}  
  \subfigure[CL after NC\label{exp_motivation:NC_cl}]{\includegraphics[width=1.8in]{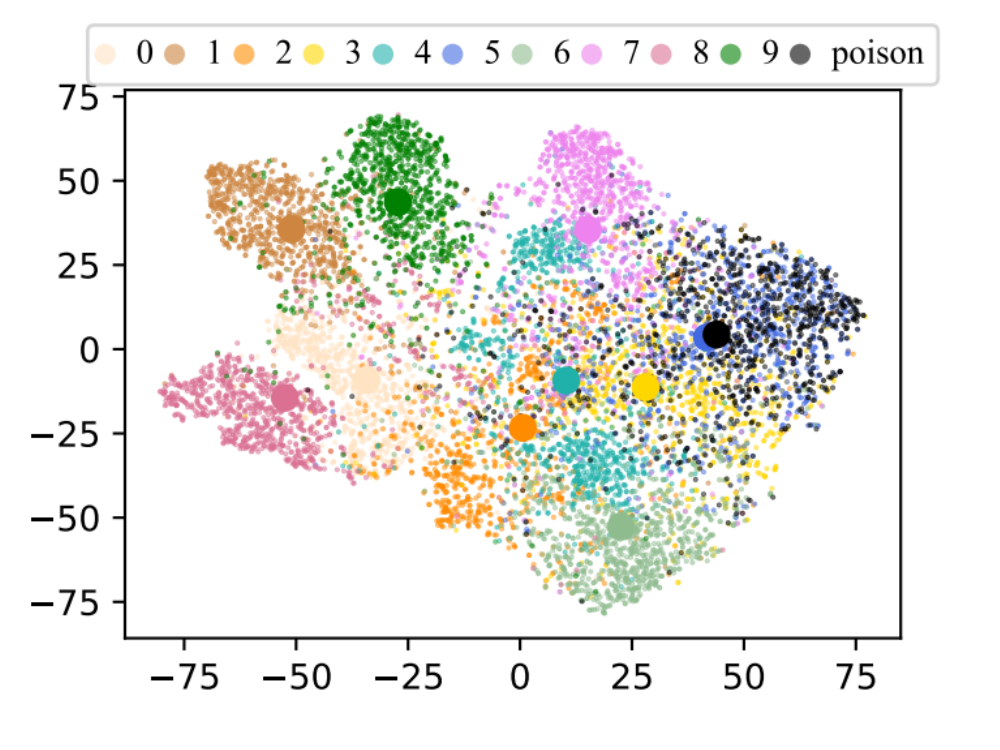}} 
   \subfigure[CL after MCLDef\label{exp_motivation:MCL_cl}]{\includegraphics[width=1.8in]{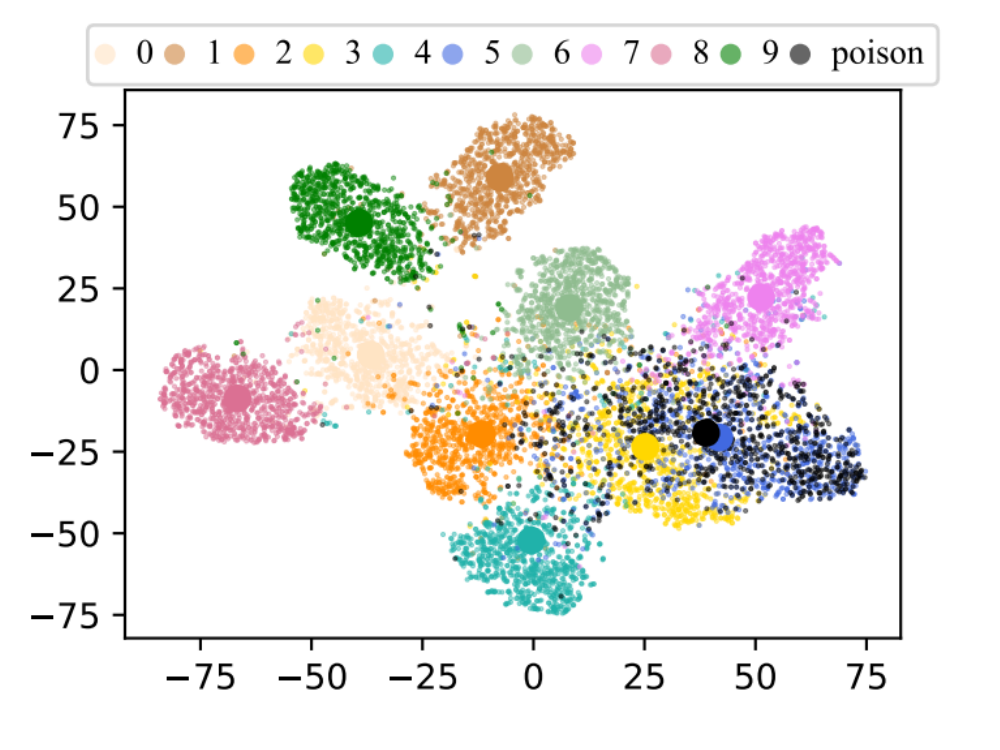}} 
  \caption{The t-SNE  of the feature representations under different attacks after defenses.} \label{exp_motivation}
\vspace{-0.05in}
\end{figure*}

\subsection{Comparison with SOTA Defense Methods (RQ1)} 
\label{section_comparison}


Table~\ref{comparison} compares the  performance  of  
MCLDef with the ones of the four SOTA defense methods  against  five 
well-known attack methods. Here, we use  {\it Before} 
to indicate the method with no defense. 
Note that to enhance the defense performance,  here we adopt the data augmentation techniques for backdoor trigger elimination for all the 
defense methods. Please refer to Appendix \ref{aug} for more results 
about the impacts of data augmentation.

From this table, we can find that for the dataset Cifar-10, 
MCLDef achieves the best ASR among all the five defense methods. 
and the best BA in four out of five attacks. 
 Specifically, compared with the second-best results marked with underscores, 
 MCLDef can reduce the ASR by 21.27\%, 32.12\%, 8.22\%, 95.79\%, and 47.72\% for the 
 five defense methods, respectively. 
 For the Sig attack, the reason of low BA  for MCLDef  (i.e., 73.81\%)
 is mainly due to the bad
 quality of  trigger inversion, which is also reflected in the case of NC (i.e., 73.34\%).
 Note that the datasets  GTSRB and ImageNet often cause the failures for 
 clean-label attacks, we do not provide their comparison results here. 
 We can observe that, for the datasets GTSRB and ImageNet,
  MCLDef can achieve the best BA and top-2 best ASR for all the cases. 
 As an example for GTSRB with Blend attack, MCLDef outperforms NC by up to 70.37\% in ASR, 
 and outperforms FT by 3.3\% in BA. 
 
 \begin{table*}[t]
\caption{Defense performance comparison of different  trigger inversion-based methods on Cifar-10.}
\centering
\begin{tabular}{c|cc|cc|cc|cc|cc}
\hline
\multirow{2}{*}{Attack} & \multicolumn{2}{c|}{Before} & \multicolumn{2}{c|}{NC} & \multicolumn{2}{c|}{MCLDef} & \multicolumn{2}{c|}{MESA} & \multicolumn{2}{c}{MCLDef(True)} \\ \cline{2-11} 
                        & ASR          & BA           & ASR        & BA         & ASR          & BA           & ASR           & BA            & ASR          & BA            \\ \hline \hline
BadNets                 & 99.64        & 83.01        & 3.58       & 81.37      & 2.22         & 82.19        & {\ul 1.34}          & {\ul 82.76}         & \textbf{0.66}         & \textbf{83.24}         \\
Trojan                 & 99.38        & 82.84        & 3.86       & 79.61      & 2.62         & 82.04        & {\ul 1.50}           & {\ul 82.76}         & \textbf{0.37}         & \textbf{83.43}         \\
Blend                   & 98.98        & 84.20         & 3.01       & 80.94      & {\ul 2.12}         & \textbf{83.98}        & /             & /             & \textbf{1.16}         & {\ul 83.97}         \\
CL                      & 95.47        & 86.44        & 5.71       & 77.31       & 0.24         & 83.97        & {\ul 0.09}          & {\ul 84.98}         & \textbf{0.02}         & \textbf{85.24}          \\
Sig                     & 99.03        & 84.64        & 17.33      & 73.34      & {\ul 3.56}         & {\ul 73.74}        & /             & /             & \textbf{0}            & \textbf{81.92}     \\ \hline   
\end{tabular}   \label{reverse_trigger}
\end{table*}
 
\subsection{Effects of MCL on Feature Space (RQ2)}

To show the reason why MCLDef outperforms other defense methods, we investigate 
the feature distributions in feature space. Due to the space limitation, we only 
provide the results for the two representatives (i.e., BadNets and CL) of  poisoned-label  attacks and clean-label 
attacks.

Based on Cifar-10, Figure \ref{exp_motivation} visualizes the  distributions in  feature space
for the two representative attacks, where each backdoored DNN is purified by three
defense methods (i.e., NAD, NC and MCLDef). 
Note that, unlike the Figure \ref{figure_motivation} that only uses 
the training data for t-SNE visualization, 
all the six subfigures in Figure \ref{exp_motivation} show the t-SNE using both 
training data and extra 5\% clean data for purification. 
From Figure \labelcref{exp_motivation:NAD_badnet,exp_motivation:NC_badnet,exp_motivation:MCL_badnet}, we can find that the clusters of  poisoned samples in different subfigures 
are all damaged by 
the three defense methods, respectively.  However, 
in Figure \labelcref{exp_motivation:MCL_badnet} there are fewer 
black features locating outside the clusters than the other two 
subfigures, indicating that MCLDef can achieve 
better BA after defending BadNets attacks. 
Similarly, in \labelcref{exp_motivation:NAD_cl,exp_motivation:NC_cl,exp_motivation:MCL_cl} the cluster of poisoned samples by CL is also 
damaged. Since  CL is a clean-label attack
that focuses on the attacking samples with a specific target label
(indicated by the blue color here), we can find that 
the black features locate in the blue cluster in a 
more concentrated manner, thus leading to  a better BA.

\subsection{Applicability Analysis (RQ3)} 
\label{applicability_of_mcl}

\begin{figure}[h]
\vspace{-0.15in}
  \centering
  \subfigure[ASR (against BadNets) \label{clean_data_rate:badnet_asr}]{\includegraphics[width=2.0in]{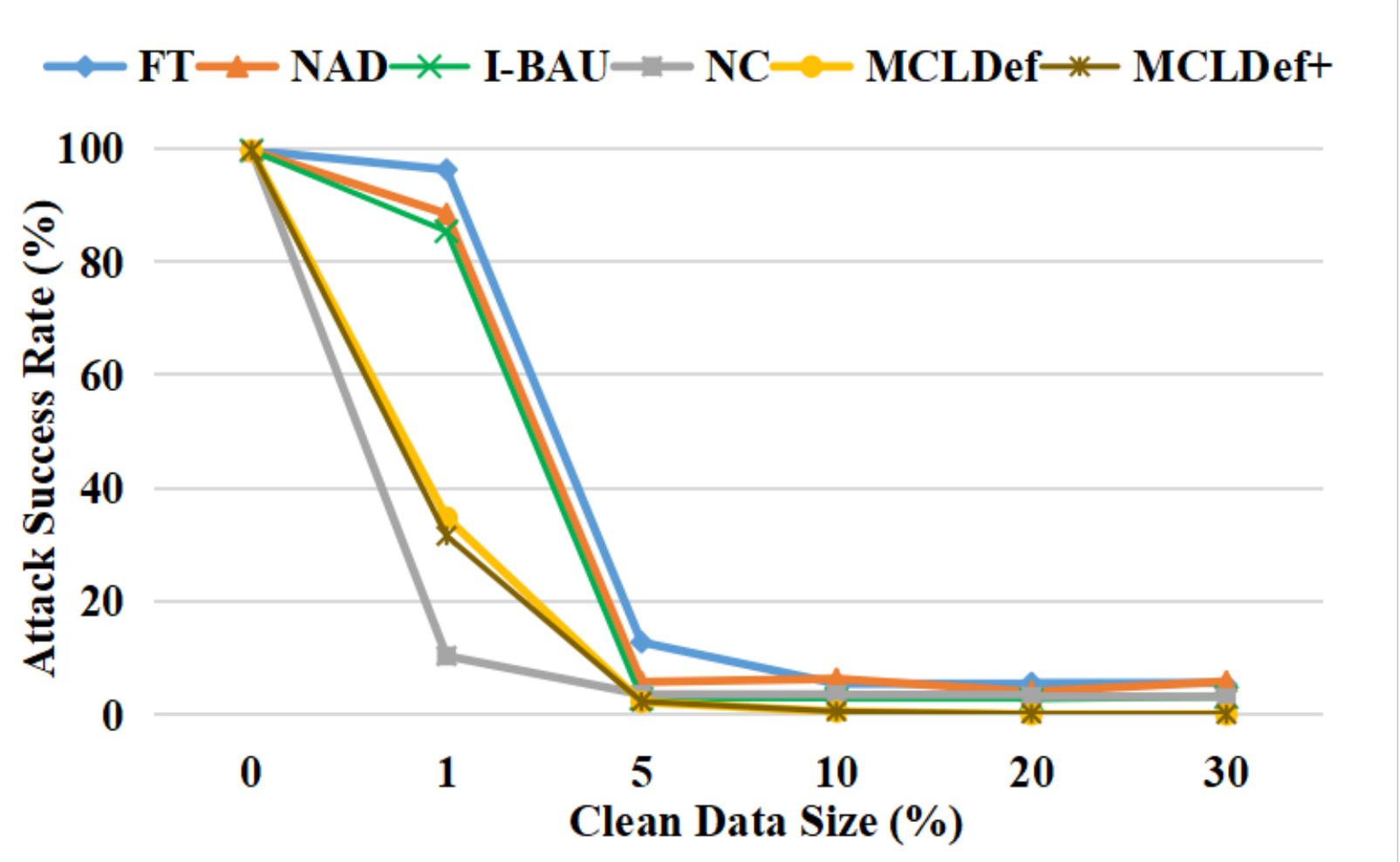}}  
  \subfigure[BA (against BadNets) \label{clean_data_rate:badnet_ba}]{\includegraphics[width=2.0in]{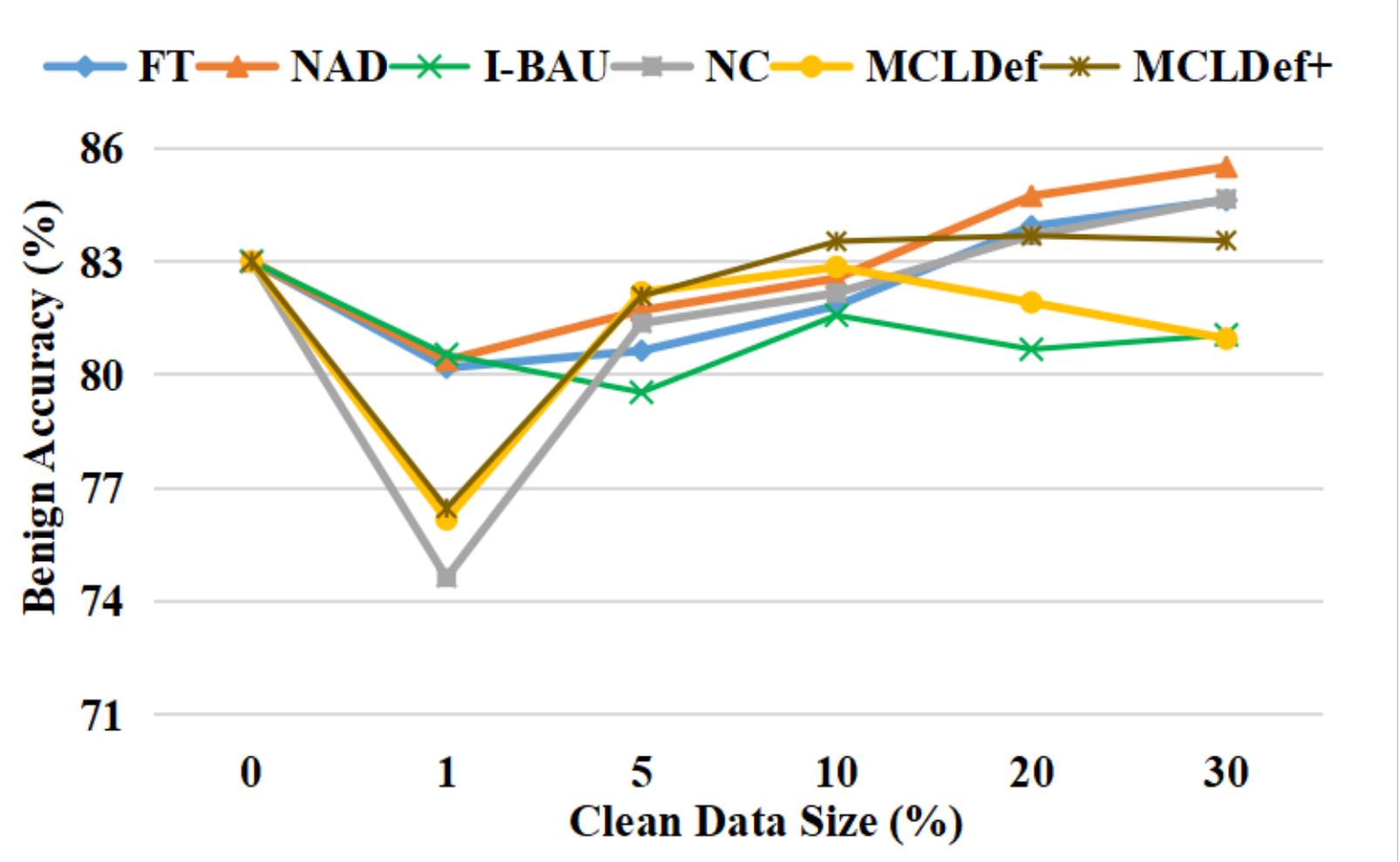}} 
  \caption{The defense performance  against BadNets with  different clean data sizes on Cifar-10.} \label{clean_data_rate}
  \vspace{-0.05in}
\end{figure}

\textbf{Impact of Clean Data Sizes.}
Based on dataset Cifar-10, Figure \ref{clean_data_rate} shows
the defense performance trends of  five defense methods (FT, NAD, I-BAU,
NC, MCLDef) against BadNets attack with different sizes of clean data. Please refer to Appendix \ref{appendix_exp_clean_data_size} 
for more results of  other four backdoor attacks. 
From this figure, we can find that more clean data will lead to lower 
ASR and higher BA for the purified DNNs of all the defense methods. Note that with 
only 1\% clean data, both NC and MCLDef can achieve significant ASR reduction, while 
 ASR improvements of the other three defense methods are negligible. 
However, in this case both NC and MCLDef need to sacrifice their BA, since the 
the quality of trigger inversion is poor due to a limited number of clean data. 
When the clean data rate is in between the range $[5\%, 10\%]$
 we can find that  ASRs of the five 
attacks will not change notably, and MCLDef can achieve the best ASR and BA. 
When the clean data rate reaches 20\%, although 
MCLDef  has the best ASR, its BA drops. Note that in this case the BA of 
FT, NAD, and NC becomes
 higher   than  the BA (i.e., 83.01\%)   of the  method without defense. 
 The main reason of the decreasing  trend of BA is because
MCLDef is based on MCL, where the distance between  two features
within a negative becomes farther along the backdoor purification process. 
When more and more clean data are involved in 
backdoor trigger elimination, 
such mutual exclusion between features imposed by 
negative pairs will more easily result in the shape change of 
benign clusters, leading to lower BA. 
To verify the reason of this phenomenon, 
we developed a new method MCLDef+ with 
one  more 
stage than MCLDef, where the third stage only 
uses the positive pairs to eliminate the backdoor from 
the purified model generated by stage two. 
The goal of the third stage is to increase
BA by further 
pulling the features in 
positive pairs closer to their corresponding benign clusters.
Please see 
Figure \ref{figure_overflow_stage3} in Appendix \ref{appendix_stage3} for more  details about MCLDef+. 
From Figure \ref{clean_data_rate:badnet_ba}, we can find
that when more clean data are provided, 
MCLDef+ can achieve better BA than MCLDef. 
Note that if the size of available clean data is small (i.e., clean sample rate is less than 10\%), 
we suggest to use MCLDef. Otherwise, MCLDef+ is a better choice  for backdoor defense.


\textbf{Influence of Trigger Inversion Quality.}
To understand the impact
of trigger inversion quality on  MCLDef, we investigated various 
trigger inversion  methods, which are used to replace the first stage of our 
approach. For example, we tried to use  Max-Entropy Staircase Approximator (MESA) \cite{qiao2019defending}
as an alternative (with $\alpha=0.1$ and $\beta=0.8$) to replace our  trigger inversion part in the first stage. 
Since MESA makes  a stronger assumption that 
both the  size and location of a trigger is known a prior,  its 
trigger inversion quality if better than ours. 
Note that due to  limitations specific in MESA, we only 
implemented the MESA-based defense against three attacks excluding  Blend and Sig attacks. 
Moreover, instead of using trigger inversion, we developed a method named 
{\it MCLDef(True)}, where the first stage is replaced with a hard-coded real trigger
with the highest trigger inversion capability among all the methods. 
Table \ref{reverse_trigger} compares the performance of all the trigger-inversion based methods. 
From this table, we can find that the higher quality the trigger inversion method has,  the better ASR and BA the corresponding defense can achieve.


\section{Conclusion}

Inspired by the observation that in the feature space of a backdoored DNN
the feature representations of both benign samples  with
the same category and the poisoned samples are separately clustered, 
we proposed a novel two-stage model-contrastive learning  framework
named MCLDef for the purpose of backdoor defense. 
Based on our proposed definitions of 
positive and negative pairs of features, 
MCLDef can shrink or even damage the cluster of poisoned samples 
and pull the poisoned samples towards the clusters of their benign counterparts. In this way, the backdoor triggers in DNNs are eliminated and 
the classification accuracy on benign samples can be guaranteed. 
Comprehensive experimental results shows that, compared with various kinds of
state-of-the-art defense methods, 
our approach can not only achieve significantly lower ASR  but also have a better benign accuracy under the help of  only 5\% clean data.

\bibliographystyle{unsrt}
\bibliography{bibfile}

\clearpage

\appendix

\section{Implementation Details for Section \ref{motivation}}    \label{appendix_reproduce_backdoor}

\begin{figure}[ht]
  \centering
  \subfigure[Badnets]{\includegraphics[width=1in]{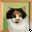}}  
  \subfigure[Trojan]{\includegraphics[width=1in]{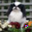}} 
   \subfigure[Blend]{\includegraphics[width=1in]{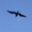}} 
   \subfigure[CL]{\includegraphics[width=1in]{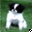}} 
   \subfigure[Sig]{\includegraphics[width=1in]{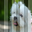}} \\
  \caption{Examples of backdoored Cifar-10 images by five different attacks.} \label{poisoned_sample}
\end{figure}

\textbf{Attack Settings.}
We evaluated the effects of five backdoor injection attacks on the Cifar-10 dataset, which can be classified into two categories: i) poisoned-label attacks including BadNets \cite{badnets}, Trojan \cite{trojann}, Blend \cite{physics_backdoor}, where the poisoned
data rate is 10\% and the target label is $y_t=5$; and ii) clean-label attacks including CL \cite{CL}, Sig \cite{Sig}, where the poisoned
data rate is 8\% and the target label is $y_t=5$. For BadNets, we used the grid trigger with a  size of  3$\times$3 placed in the bottom 
right corner of the image. For Trojan, we used a 3$\times$3 trigger placed in the bottom 
right corner of the image. For Blend, to better reflect the advantages of our defense method, we used a random color trigger with a size of 3$\times$3 placed in the center of the image, and the blend rate was set to 0.1.
For CL, we followed the same settings as used in \cite{CL}, where  
we used the Projected Gradient Descent (PGD) \cite{micikevicius2017mixed} under the $l_\infty$ norm with a maximum perturbation size $\epsilon=8$ to generate adversarial perturbed data, and adopted a grid trigger with a  size
of 3$\times$3 placed in the bottom 
right corner of the image. For Sig, we generated the trigger by following the horizontal sine function as defined in \cite{Sig}, where  $\Delta=20$ and $f=6$. Figure \ref{poisoned_sample} gives
the examples  generated by 
the five backdoor attacks in our experiments.

\textbf{Training Setting.}
To train a poisoned model  on Cifar-10 for the examples shown in Figure \ref{figure_motivation}, we set its SGD optimizer with a momentum of 0.9, a weight decay of $1\times10^{-4}$, and an initial learning rate of 0.1. With a batch size of 64, we trained the WRN-16-1 network for 50 epochs. Note that at the $20^{th}$ and the $35^{th}$ epoch, the learning rate will be reduced by a factor of 10, respectively. To avoid side effects of backdoor attacks, we did not use any data augmentation techniques.

\textbf{t-SNE Visualization Setting.}
To observe the distribution characteristics of both benign and poisoned  samples in feature space, we used different colors in t-SNE visualization for  benign samples with a specific category and  poisoned samples, respectively. 
Here, we use the feature vectors obtained from the 
 feature extractor (a DNN without  the last layer of classifiers) as the inputs of t-SNE \cite{van2008visualizing} for visualization. 
 To enable  better visualization effect, 
 all the t-SNE figures in this paper show all the 
 investigated poisoned samples and partial (i.e., 20\%) clean samples  
 that are randomly selected from each category of a dataset.

\section{Implementation Details for Section \ref{exps}}     \label{appendix_main_exp}

\subsection{Settings of Datasets  and DNNs}   
\label{appendix_dataset}
Table \ref{dataset} presents the statistics of datasets and DNNs used in our experiments.
\begin{table*}[ht]
\caption{Statistics of datasets}
\centering
\setlength{\tabcolsep}{1mm}{
\begin{tabular}{c c c c c c}
\hline
Dataset  & Input Size  & Classes & Trainging Images & Test Images & DNN model\\ \hline \hline
Cifar-10 & 3 $\times$ 32 $\times$ 32 & 10      & 50000            & 10000  & WideResNet-16-1     \\ 
GTSRB    & 3 $\times$ 32 $\times$ 32 & 43      & 26640            & 12569  & WideResNet-16-1     \\
ImageNet subset    & 3 $\times$ 224 $\times$ 224 & 20      & 26000            & 1000   & ResNet-18    \\ \hline
\end{tabular}} \label{dataset}
\end{table*}

\subsection{Settings of Attacks} \label{appendix_attack}

\begin{figure}[ht]
  \centering
  \subfigure[Badnets]{\includegraphics[width=1in]{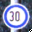}}  
  \subfigure[Trojan]{\includegraphics[width=1in]{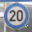}} 
   \subfigure[Blend]{\includegraphics[width=1in]{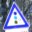}} 
  \caption{Examples of backdoored GTSRB images by three different attacks.} \label{poisoned_sample_gtsrb}
\end{figure}

\begin{figure}[ht]
  \centering
  \subfigure[Badnets]{\includegraphics[width=1in]{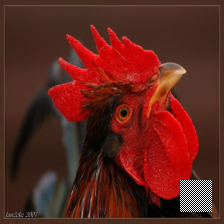}}  
  \subfigure[Trojan]{\includegraphics[width=1in]{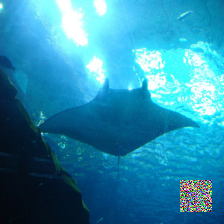}} 
   \subfigure[Blend]{\includegraphics[width=1in]{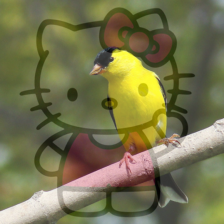}} 
  \caption{Examples of backdoored ImageNet images by three different attacks.} \label{poisoned_sample_ImageNet}
\end{figure}

\textbf{Attack Details.}
We applied  five backdoor injection attacks on the Cifar-10 dataset \cite{krizhevsky2009learning}, which can be classified into two categories: i) poisoned-label attacks including BadNets \cite{badnets}, Trojan \cite{trojann}, Blend \cite{physics_backdoor}, where the poisoning data rate is 10\% and the target label is $y_t=5$; and ii) clean-label attacks including CL \cite{CL}, Sig \cite{Sig}. The attack settings for Cifar-10 are the same  as the ones described in Appendix \ref{appendix_reproduce_backdoor}. 
Note that we only performed the poisoned-label backdoor attacks
(i.e., BadNets, Trojan, and Blend) on GTSRB dataset \cite{stallkamp2012man}. We imposed the triggers on  GTSRB images 
in a same way as the one used in Cifar-10 as described in 
Section \ref{appendix_reproduce_backdoor}. 
Figure \ref{poisoned_sample_gtsrb} presents
three examples of  backdoored   GTSRB images.
We only conducted the poisoned-label backdoor attacks on 
the ImageNet subset \cite{deng2009ImageNet},  where the 
poisoned data rate is 10\% and the   target label is $y_t=5$. 
For BadNets and Trojan attacks on  ImageNet images, we injected 
a grid trigger with a  size of  32$\times$32 in 
the bottom 
right corner of the image. For Blend attacks, we applied 
the ``Hello Kitty''  trigger on ImageNet images. Figure \ref{poisoned_sample_ImageNet} presents
three examples of  backdoored   ImageNet images.

\textbf{Training Settings.}
For datasets Cifar10 and GTSRB, we used the same training settings
as  those described in Appendix \ref{appendix_reproduce_backdoor}. 
Note that for    GTSRB, we set the initial learning rate of the optimizer to 0.01. 
 For the  ImageNet subset, we trained the Resnet-18 network for  100 epochs with a batch size of 32, where  the learning rate was reduced by a factor of 10 at 
  the $20^{th}$ and the $70^{th}$ epoch,  respectively. 
 To enhance the generalization of target DNNs, we used various
 data augmentation techniques, e.g., random crop and random horizontal flipping. All the other settings are the same as those applied on  Cifar-10.


\subsection{Settings of Defenses} 
\label{appendix_defense}

For the purpose of trigger inversion, in MCLDef and NC
we optimized the trigger pattern and mask using the Adam optimizer with a momentum of 0.9, and a learning rate of 0.005. 
The optimization involves 100 epochs with a batch size of 64.
For datasets Cifar-10 and GTSRB, we set $\lambda$ of the
 optimization objective to 0.01. For  the ImageNet subset, 
 we set  $\lambda=0.001$.

For the backdoor elimination purpose, 
 we finetuned all the backdoored model  (e.g., the purified model in MCLDef)
for 20 epochs for all the five defense methods, where the  
learning rate was reduced by a factor of 10 at both  the  
$2^{nd}$ and  $10^{th}$ epochs.  Note that 
to enhance the defense quality, we used  data augmentation techniques (including random crop and random horizontal flipping) for finetuing. 
 For  NAD, when dealing with datasets Cifar-10 and GTSRB, we use the same setting as described in \cite{li2021neural}. 
 When dealing with the ImageNet subset, 
 we calculated the NAD loss only based on 
  the last three residual groups of  ResNet-18. 
  To achieve a best trade-off between ASR and BA, we 
  set $\beta=5000$.
 Since  the settings of MCLDef used in this paper
 are not suitable for I-BAU, in the experiment  I-BAU used
 the settings as described in \cite{li2021anti}.

Based on the above defense settings, we implemented the defense methods FT and NAD by modifing the source code of 
NAD\footnote{\url{https://github.com/bboylyg/NAD}}. 
We adopted  NC  based on the open source code from its official website\footnote{\url{https://github.com/bolunwang/backdoor}}. 
By slightly modifying the part of dataset loading and model loading, 
we implemented I-BAU based on the open-source code from its 
official website\footnote{\url{https://github.com/YiZeng623/I-BAU}}.


\section{Supplementary Experimental Results}     \label{appendix_more_exps}
In this section, we provide
more experiments to show the advantages of  MCLDef, whose  code is available at \url{https://github.com/WeCanShow/MCL}.

\subsection{Impact of Data Augmentation Techniques}
\label{aug}

To investigate the impact of data augmentation, 
here we considered two cases:
i) data augmentation is not used in backdoor trigger elimination; and ii) data augmentation is used in the training of poisoned model.
Here, we only consider two kinds of data augmentation techniques, random crop and random horizontal flipping.

\textbf{Defense without Data Augmentation.}
Table \ref{defense_no_aug} shows the defense performance without
using data augmentation techniques 
for the five defense methods by using Cifar-10 dataset.  
From this table, we can find that our MCLDef can achieve the best performance in ASR, and the best BA in three out of five attacks. 
The difference between the defense performance of MCLDef in  Table \ref{comparison} and Table \ref{defense_no_aug} is negligible. 
Note that, we can find that FT and NAD methods greatly rely on 
the use of data augmentation techniques. Compared with the results in 
Table \ref{comparison}, the ASR of NAD against Trojan is 91.08\% without considering data augmentation, while   the ASR of NAD against Trojan is 5.14\% by taking the data augmentation into account.

\begin{table*}[t]
\caption{Performance comparison  of five backdoor defense methods  on Cifar-10, where  data augmentation techniques are not used in the defense.}
\centering
\setlength{\tabcolsep}{0.8mm}{
\begin{tabular}{c|cc|cc|cc|cc|cc|cc}
\hline
\multirow{2}{*}{Attack} & \multicolumn{2}{c|}{Before} & \multicolumn{2}{c|}{FT}    & \multicolumn{2}{c|}{NAD} & \multicolumn{2}{c|}{I-BAU} & \multicolumn{2}{c|}{NC} & \multicolumn{2}{c}{MCLDef}         \\ \cline{2-13} 
                        & ASR          & BA           & ASR          & BA          & ASR    & BA              & ASR    & BA              & ASR           & BA      & ASR           & BA             \\ \hline \hline
BadNets                 & 99.64        & 83.01        & 89.96        & 80.1        & 49.88  & {\ul 81.73}     & {\ul 4.12}  & 79.91     & 4.62    & 79.64   & \textbf{2.10}  & \textbf{81.81} \\
Trojan                 & 99.38        & 82.84        & 98.06        & 80.49       & 91.08  & {\ul 81.84}     & {\ul 4.32}  & 79.89     & 4.43    & 80.69   & \textbf{1.83} & \textbf{82.61} \\
Blend                   & 98.98        & 84.20        & 4.47 & 80.79       & 14.20   & {\ul 83.03}      & {\ul 2.44}  & 82.11     & 2.93          & 80.76   & \textbf{1.44}    & \textbf{84.89} \\
CL                      & 95.47        & 86.44        & 23.21        & 81.5  & 22.66   & 82.18  & 38.47  & {\ul 82.83}     & {\ul 8.42}    & 77.96   & \textbf{0.24} & \textbf{83.57}             \\
Sig                     & 99.03        & 84.64        & 96.67        & {\ul 80.09} & 84.32  & \textbf{81.43}  & 75.39  & 81.27     & {\ul 50.39}   & 71.94   & \textbf{1.66} & 72.93     \\ \hline    
\end{tabular}}   \label{defense_no_aug}
\end{table*}

\begin{table*}[t]
\caption{Performance comparison of five backdoor defense methods  on Cifar-10,  where data augmentation techniques are used in the  training training of poisoned models.}
\centering
\setlength{\tabcolsep}{0.8mm}{
\begin{tabular}{c|cc|cc|cc|cc|cc|cc}
\hline
\multirow{2}{*}{Attack} & \multicolumn{2}{c|}{Before} & \multicolumn{2}{c|}{FT}  & \multicolumn{2}{c|}{NAD}       & \multicolumn{2}{c|}{I-BAU}  & \multicolumn{2}{c|}{NC} & \multicolumn{2}{c}{MCLDef}         \\ \cline{2-13} 
                        & ASR          & BA           & ASR        & BA          & ASR           & BA             & ASR           & BA      & ASR           & BA     & ASR           & BA             \\ \hline \hline
BadNets                 & 100          & 89.39        & 4.93       & 83.09       & 9.66          & 84.81    & {\ul 1.50}          & {\ul 85.39}    & 3.10     & 81.39   & \textbf{0.02} & \textbf{89.01} \\
Trojan                 & 99.78        & 90.09        & 3.36       & 81.69       & 3.19          & 83.64    & {\ul 1.61}          & {\ul 85.88}    & 2.52    & 80.97   & \textbf{1.11} & \textbf{87.00}    \\
Blend                   & 97.40         & 89.16        & 3.24       & 82.54       & 4.23          & 85.04    & 21.58          & {\ul 85.28}    & {\ul 2.81}    & 81.40    & \textbf{2.04} & \textbf{87.06} \\
CL                      & 98.18        & 90.56        & 6.40        & 81.71       & 7.74          & 82.90     & {\ul 2.71}          & {\ul 86.94}    & 5.77    & 76.82   & \textbf{0.09} & \textbf{87.79} \\
Sig                     & 97.87        & 91.06        & 2.59 & 82.86 & \textbf{1.04} & {\ul 84.89} & {\ul 1.36}          & \textbf{85.80}    & 26.43         & 68.90    & 7.92          & 73.27    \\ \hline     
\end{tabular}}   \label{attack_aug}
\end{table*}

\textbf{Attack with Data Augmentation.}
To enhance the attack performance,  
data augmentation techniques are widely used in 
the training of poisoned models. 
To evaluate the performance of MCLDef for this case, we 
use the data augmentation techniques (i.e., random crop and random horizontal flipping) to derive training samples for poisoned models. 
To accommodate such extra training samples, in CL attack we changed the  position of triggers accordingly, i.e., we placed the trigger in the center of  images. To obtain the poisoned models for the five 
attacks, we trained each backdoored DNN for 100 epochs with an  
 initial learning rate of 0.1, which is reduced by a factor of 10 at the $20^{th}$ and $70^{th}$ epochs, respectively.  
 The other training configurations are the same as the ones used in 
 Appendix \ref{exps}.
 Table \ref{attack_aug} shows the  performance of five defense methods on the generated poisoned models. 
From this table, we can find that MCLDef can achieve the best ASR and BA for all the poisoned-label attacks. However,  although MCLDef outperforms NC for Sig in terms of both ASR and BA, its defense is not 
as good as the ones of the other three defense methods (i.e., FT, NAD, I-BAU). Based on the observations in 
Table~\ref{reverse_trigger}, the  reason 
for the low performance here is mainly due to the 
quality of trigger inversion. If a new trigger inversion method 
with higher
quality can be integrated into the first stage, the defense performance  of 
MCLDef will be greatly improved for the Sig attack.

\subsection{MCLDef+} \label{appendix_stage3}

\begin{figure}[ht]
\centering
\includegraphics[width=3in]{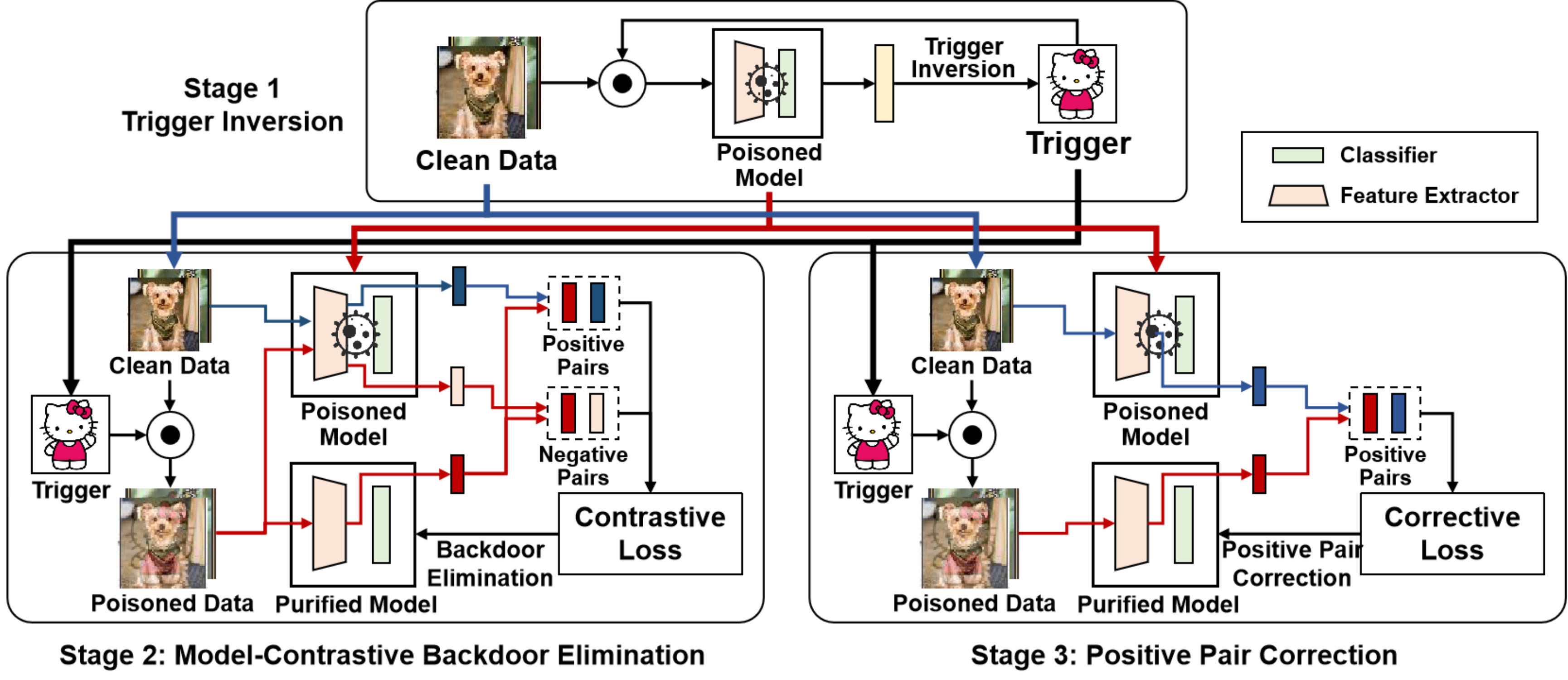}
\caption{Overview of our MCLDef+ defense method} \label{figure_overflow_stage3}
\end{figure}

As aforementioned in Section \ref{applicability_of_mcl}, we find that when the clean data rate exceeds 20\%, MCLDef has the best ASR but its BA drops. The main reason for this trend is because the distance between two features within a negative becomes farther along the backdoor purification process. When more and more clean data are involved in backdoor trigger elimination, such mutual exclusion between features imposed by negative pairs will more easily result in the shape change of benign clusters. Therefore, we add an extra stage to verify the phenomenon, named {\it positive pair correction}. Figure \ref{figure_overflow_stage3} shows an overview of  MCLDef+, where   stages 1-2 are the same as the ones of MCLDef, while stage 3 tries to further
pull the feature representations of poisoned data towards those of their benign counterparts.

Assume that  $\mathcal{D}$ is a clean dataset  with N  samples    and
the trigger obtained in  stage 1 is a combination of $\boldsymbol{\Delta^{*}}$ 
and  $\boldsymbol{m^{*}}$. 
Based on these two things, we can generate a poisoned dataset $\mathcal{\hat{D}}$ by using Equation \ref{eq_poi}. 
Let $(\boldsymbol{x_i}, y_i) \in \mathcal{D}$ be a clean sample and 
$(\boldsymbol{\hat{x_i}}, \hat{y_i}) \in \mathcal{\hat{D}}$ be its corresponding poisoned sample. 
According to the definition of positive pairs in Section \ref{construct_pair}, we can generate the corresponding positive pair $(\boldsymbol{z_i},\boldsymbol{z_i^{cle}})$ for   $\boldsymbol{x_i}$ and
$\boldsymbol{\hat{x_i}}$. 
We define our  {\it corrective loss} for MCLDef+  as follows:
\begin{equation}
    \mathcal{L}_{corrective}=\sum_{i=1}^{N}{-sim(\boldsymbol{z_i},\boldsymbol{z_i^{cle}})},
\end{equation}
where  $sim(\cdot,\cdot)$ represents the similarity of two feature vectors based on the cosine similarity function. In  stage 3, we optimize $\boldsymbol{\omega}$  as follows:
\begin{equation}
    \boldsymbol{\omega^{*}}={\underset {\boldsymbol{\omega}} { \operatorname {arg\,min} } \, \mathcal{L}_{corrective} }.
\end{equation}

\subsection{Impact of Clean Data Sizes} \label{appendix_exp_clean_data_size}
\begin{figure}[h]
  \centering
  \subfigure[ASR (against Trojan) \label{appendix_clean_data_size:trojan_asr}]{\includegraphics[width=1.3in]{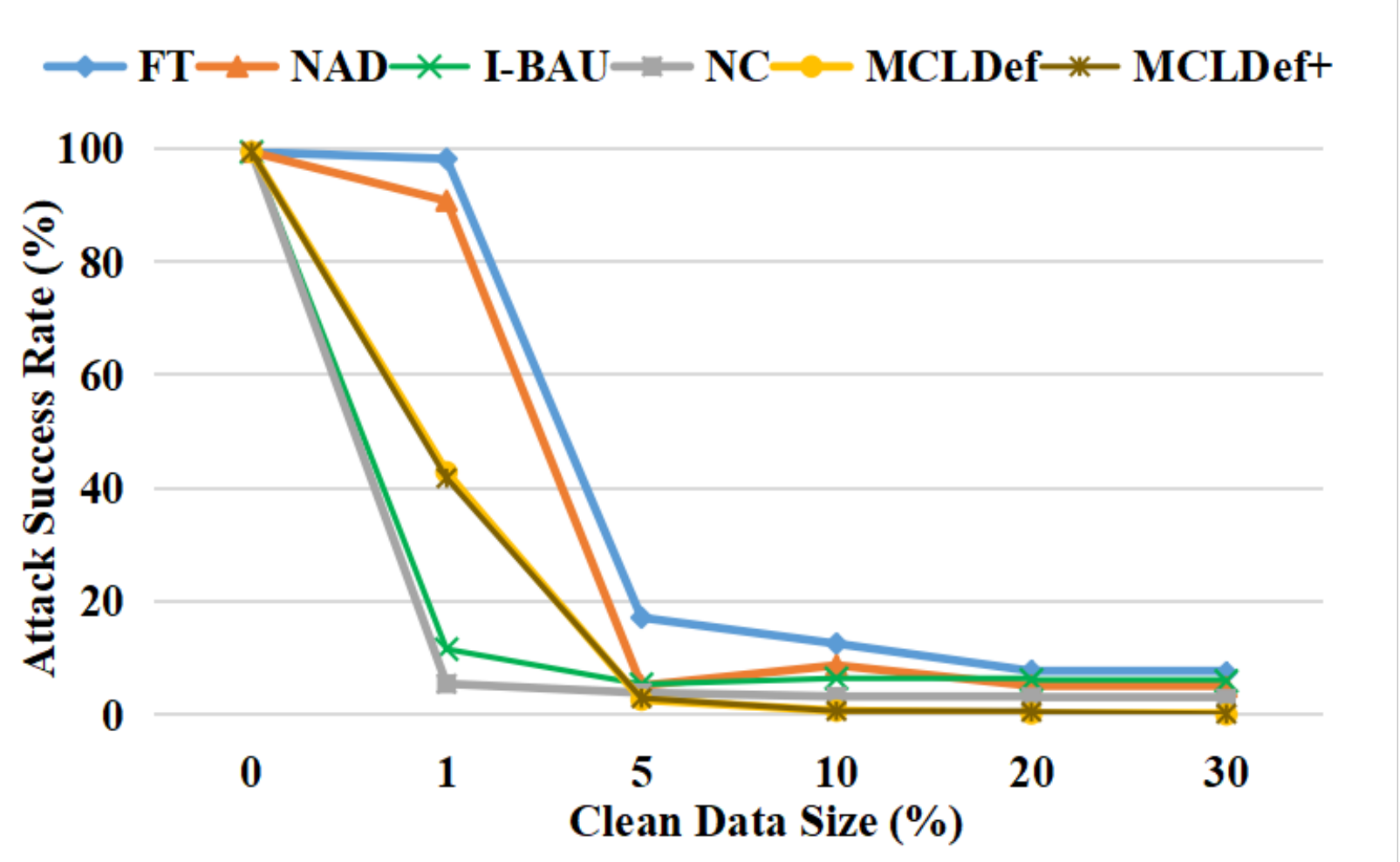}}  
  \subfigure[ASR (against Blend)\label{appendix_clean_data_size:blend_asr}]{\includegraphics[width=1.3in]{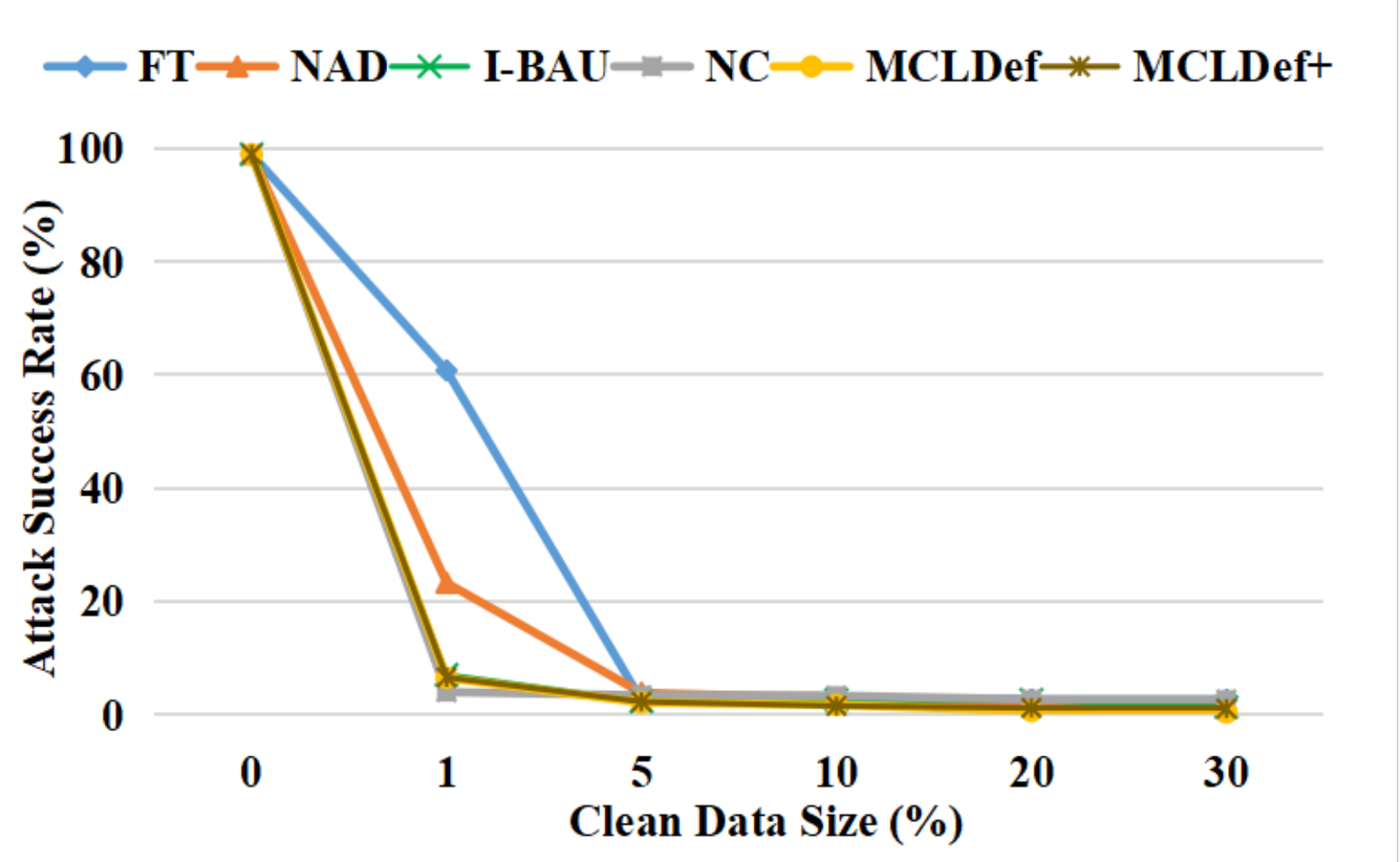}} 
   \subfigure[ASR (against CL)\label{appendix_clean_data_size:cl_asr}]{\includegraphics[width=1.3in]{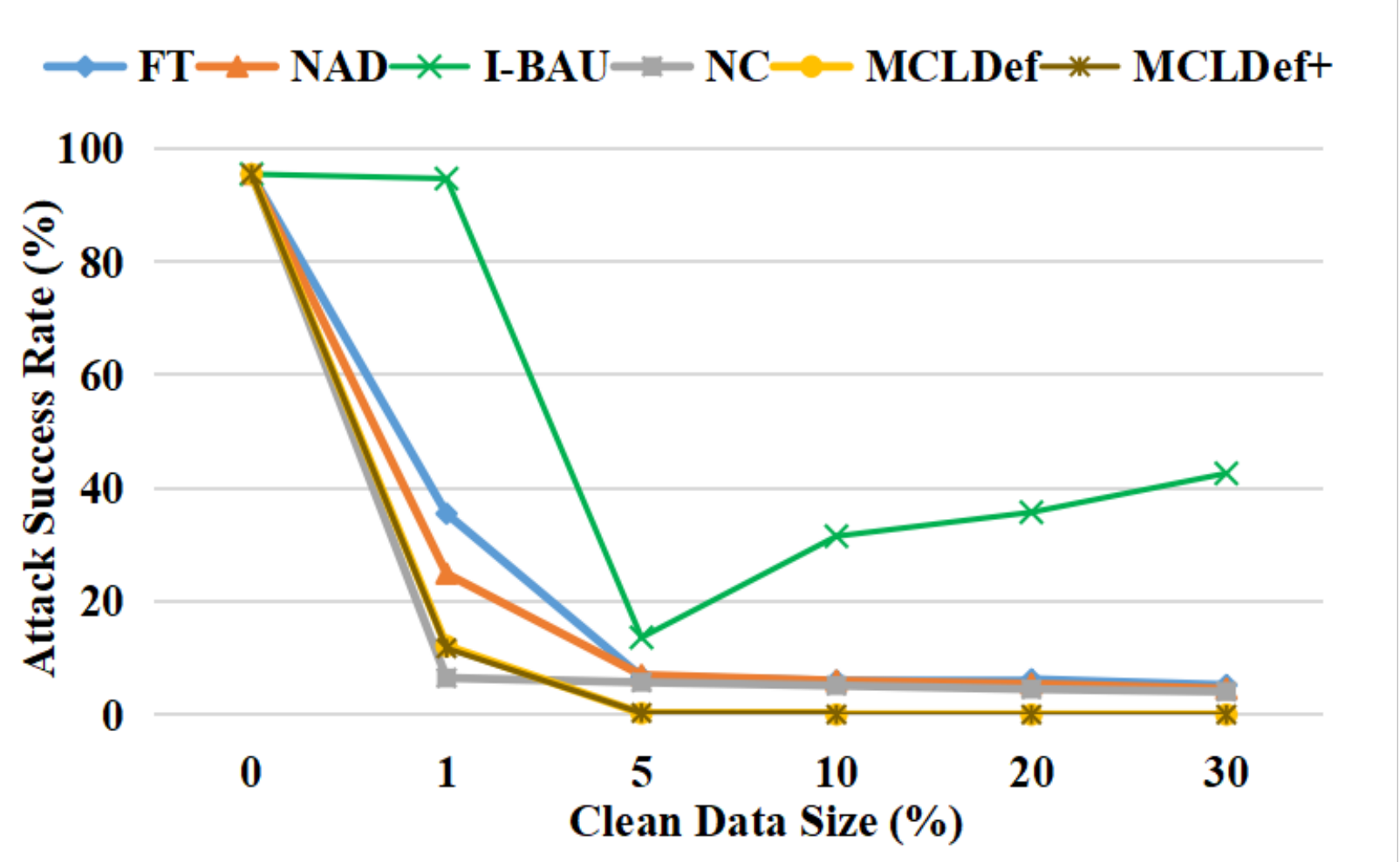}}
   \subfigure[ASR (against Sig)\label{appendix_clean_data_size:sig_asr}]{\includegraphics[width=1.3in]{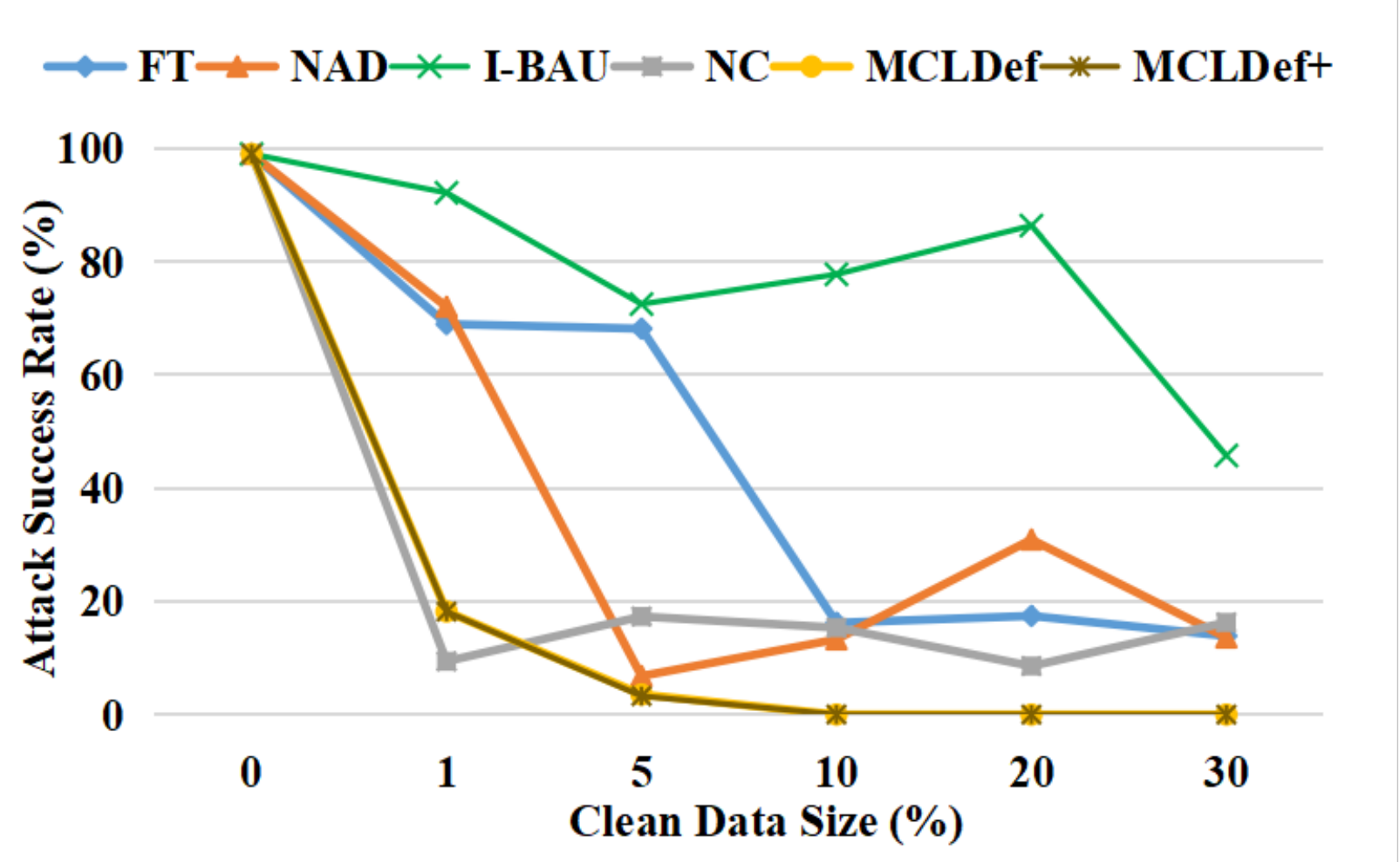}} \\
    \subfigure[BA (against Trojan) \label{appendix_clean_data_size:trojan_ba}]{\includegraphics[width=1.3in]{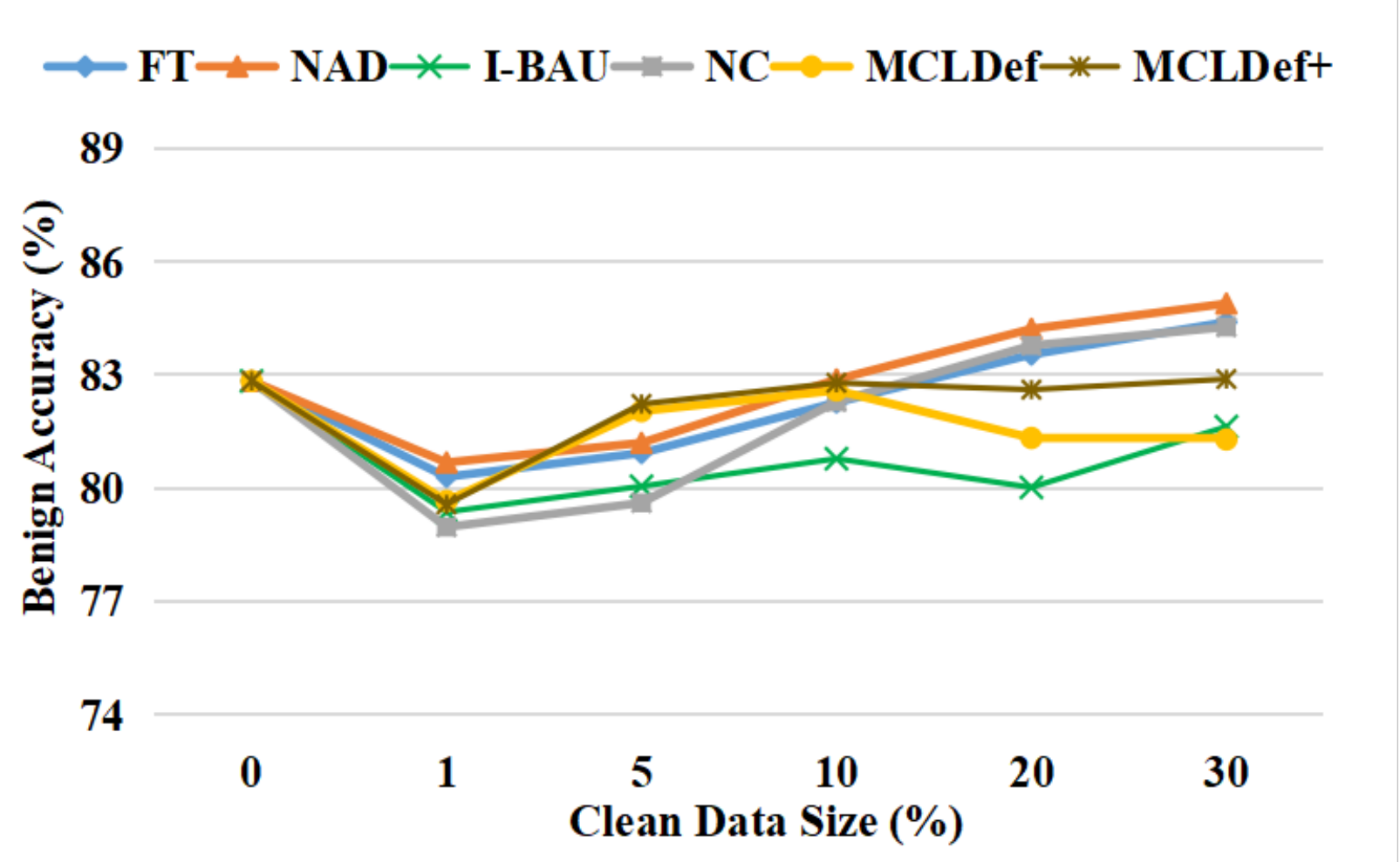}}  
  \subfigure[BA (against Blend)\label{appendix_clean_data_size:blend_ba}]{\includegraphics[width=1.3in]{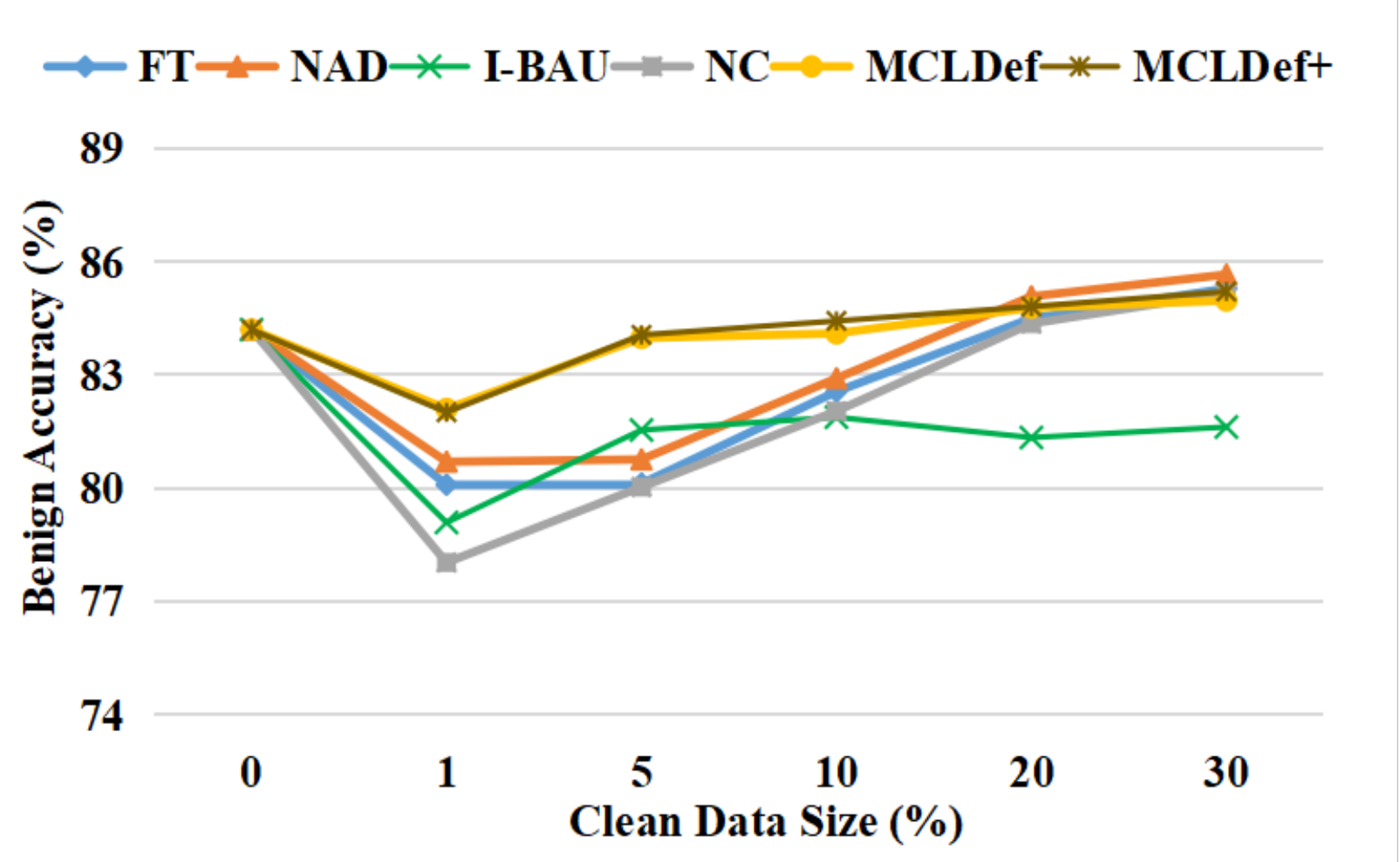}} 
   \subfigure[BA (against CL)\label{appendix_clean_data_size:cl_ba}]{\includegraphics[width=1.3in]{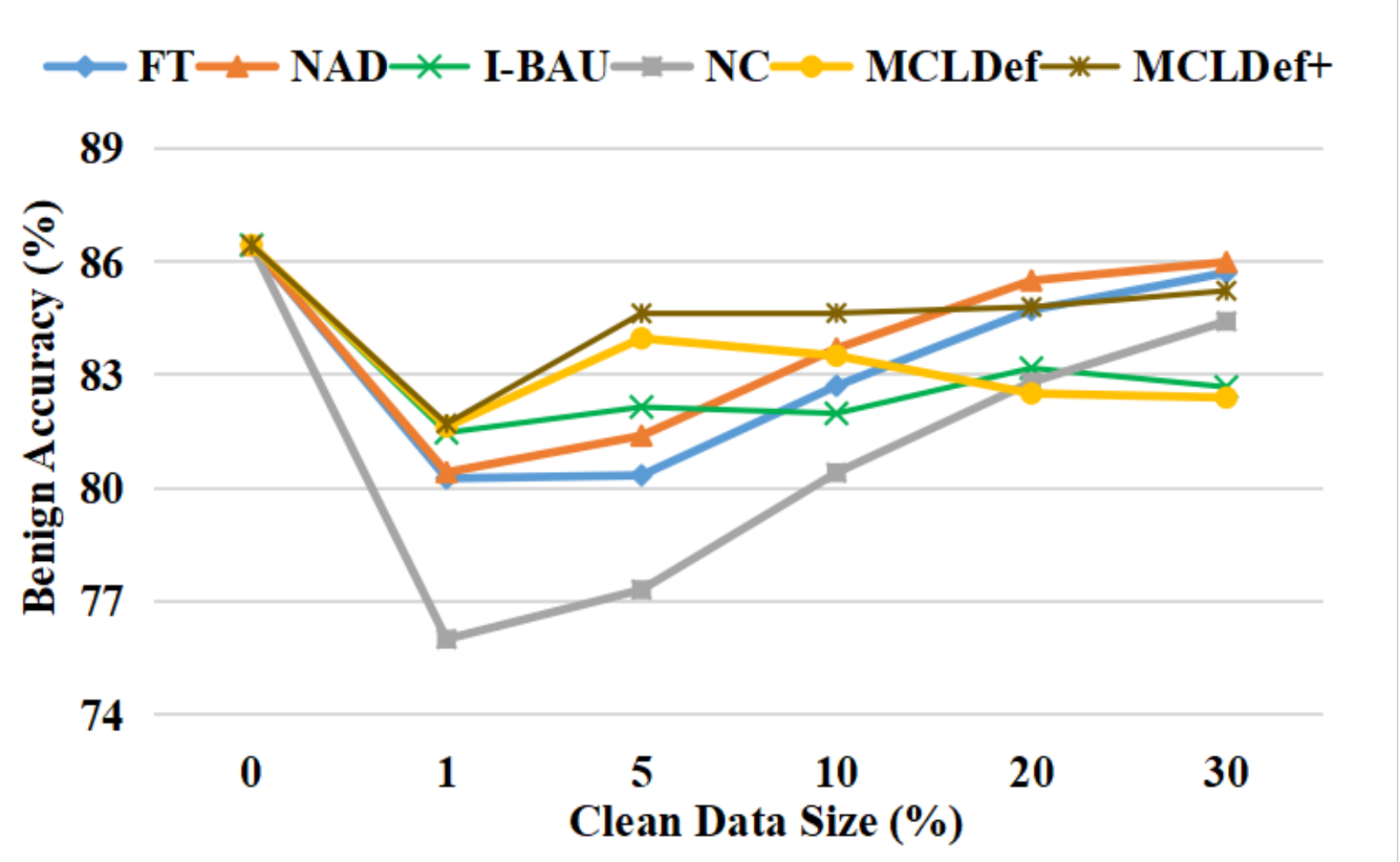}}
   \subfigure[BA (against Sig)\label{appendix_clean_data_size:sig_ba}]{\includegraphics[width=1.3in]{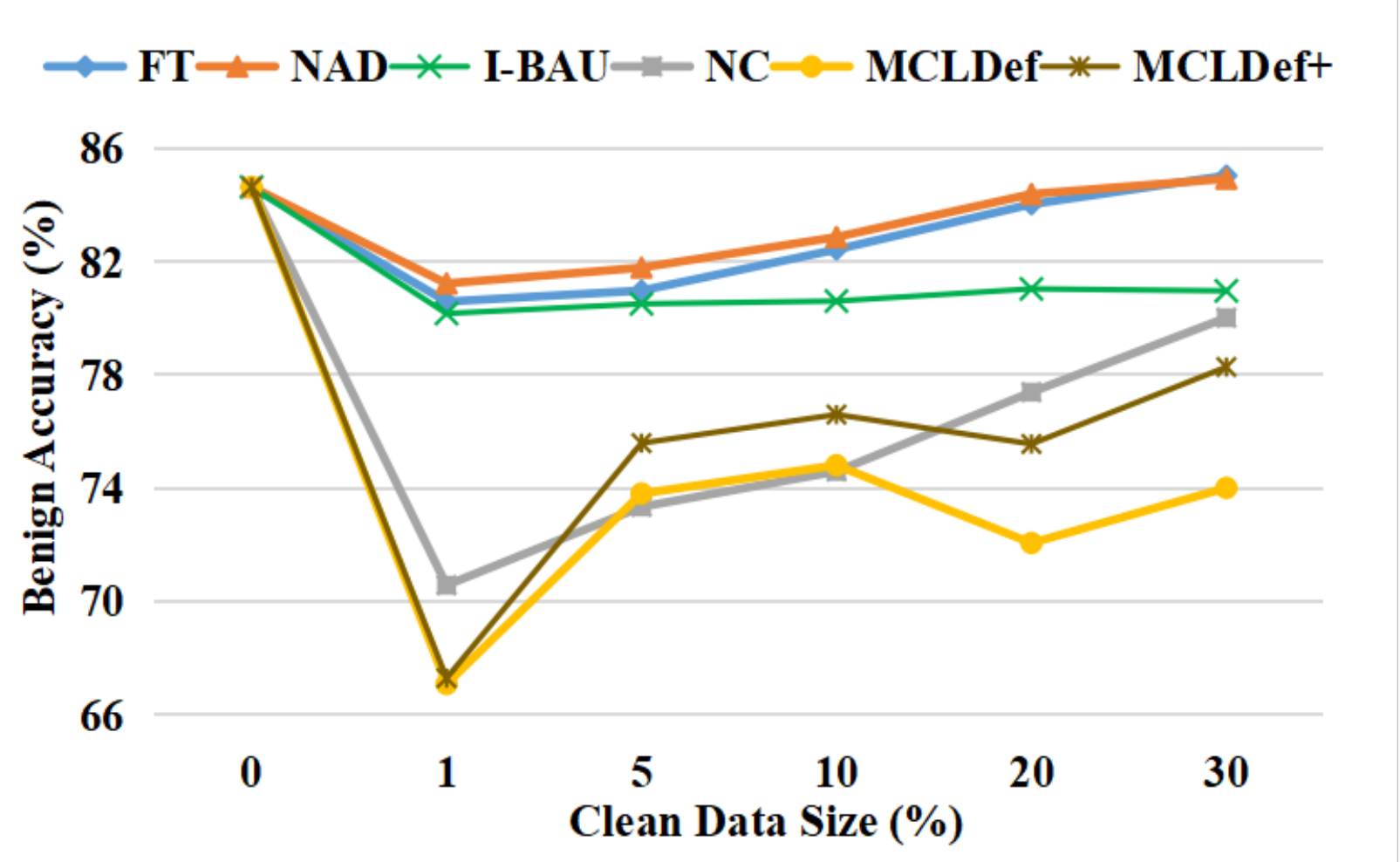}} \\
  \caption{Defense performance of five backdoor elimination
  methods under different  clean data rates.} \label{fig_appendix_exp_clean_data_size}
\end{figure}

Based on Cifar-10 dataset, Figure \ref{fig_appendix_exp_clean_data_size}
 compares
the defense performance of five defense methods against Trojan, Blend, CL and Sig attacks using 
different clean data sizes. From this figure, we can draw
the same 
conclusion as the one made in 
Section \ref{applicability_of_mcl}.
We can find that more clean data will lead to better ASR, and MCLDef and MCLDef+ can always have the best ASR.
Moreover, MCLDef+ can indeed improves the BA of  MCLDef.


\subsection{Impact of Hyperparameters}
\begin{figure}[h]
  \centering
  \subfigure{\includegraphics[width=2.3in]{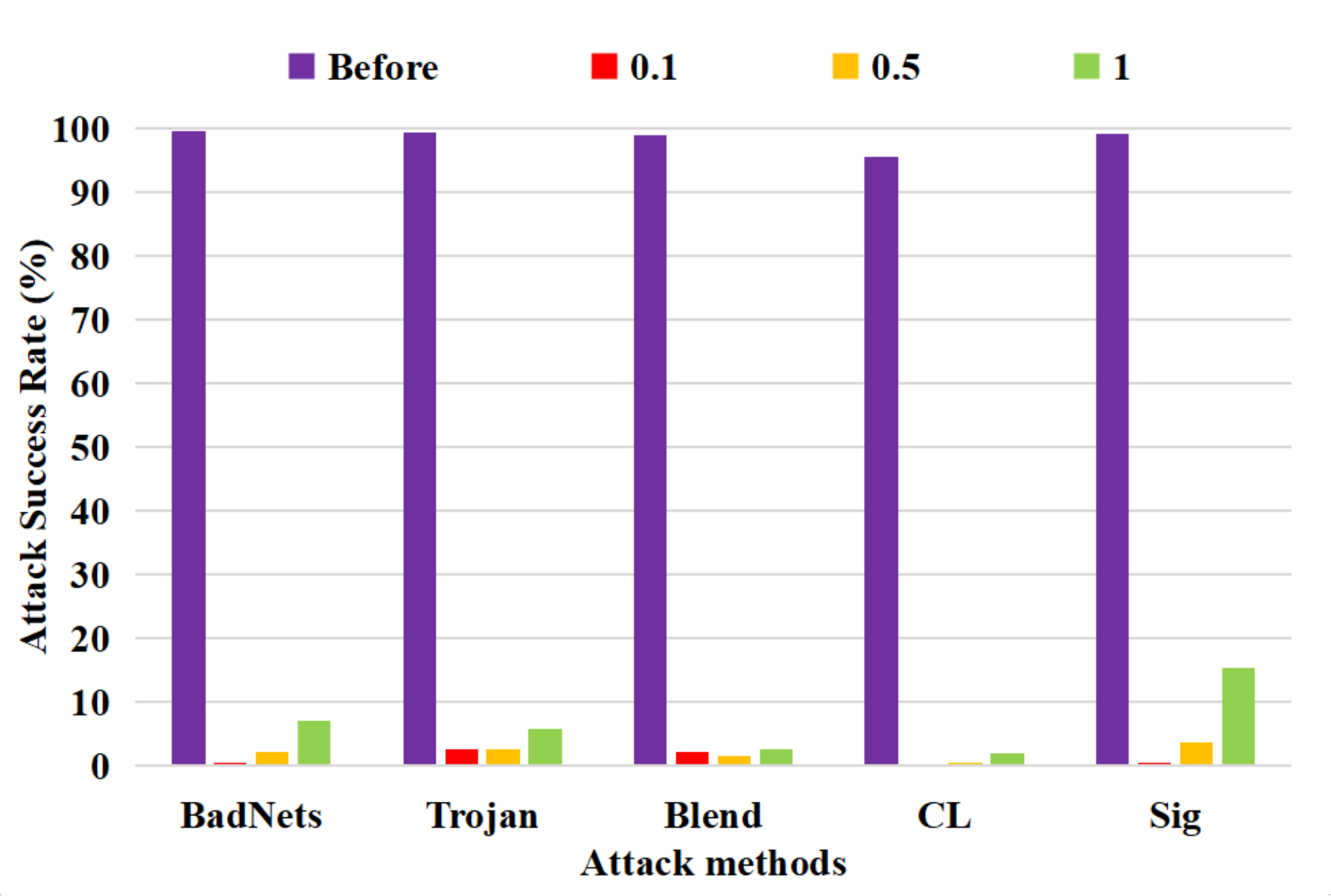}}  
  \subfigure{\includegraphics[width=2.3in]{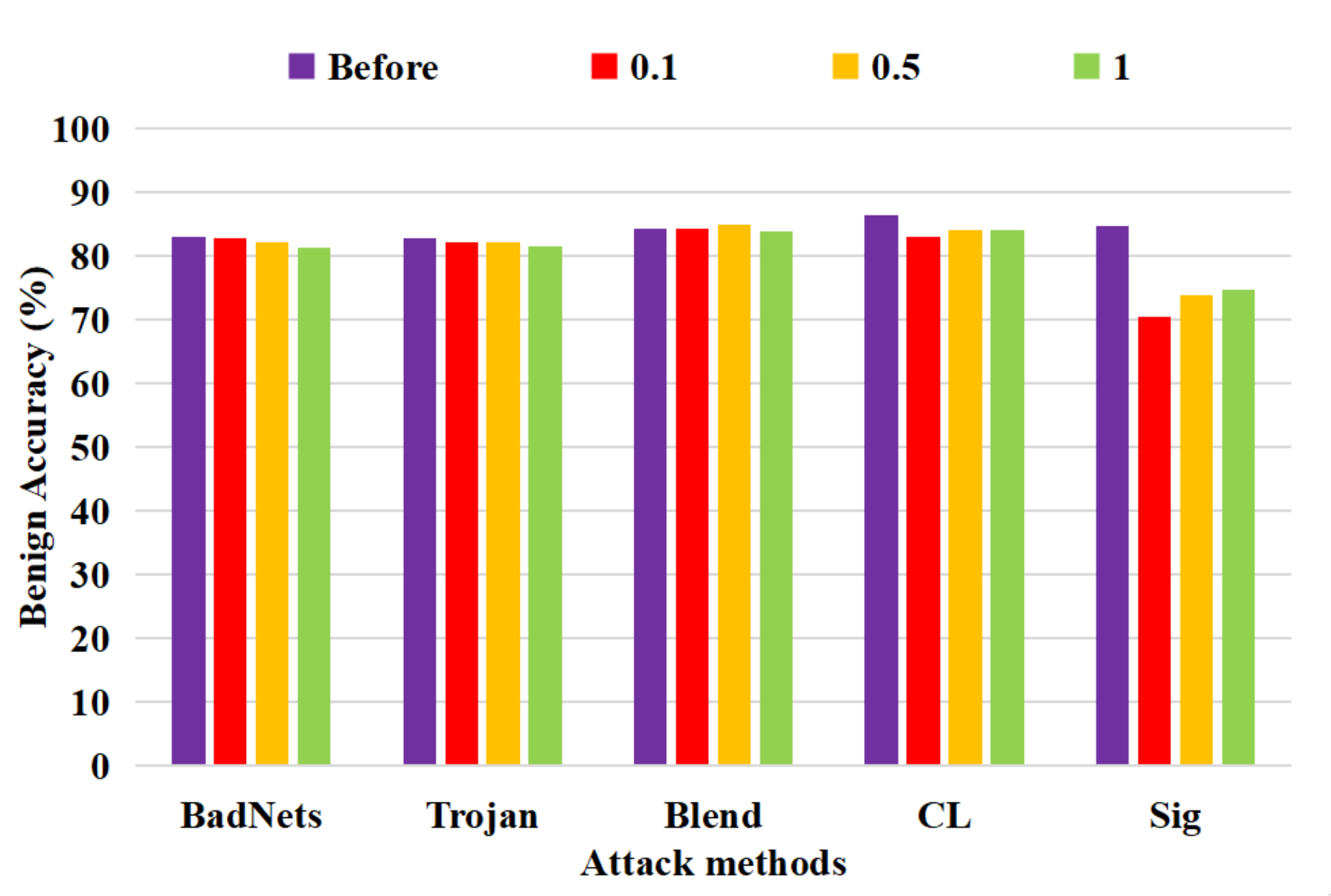}} 
  \caption{Defense performance  of MCLDef  on Cifar-10 with different $\tau$.} \label{different_tau}
\end{figure}

We considered  the impact of temperature parameter (i.e., $\tau$ in Equation \ref{contrastive_loss})
on  ASR and BA of MCLDef for the five attack methods.
Figure \ref{different_tau} shows the results with different values of 
$\tau$, where $\tau\in \left\{0.1, 0.5, 1.0\right\}$. Note that by defalt in MCLDef $\tau=0.5$.
From this figure, we can find that a smaller $\tau$ will lead to a better ASR, but may result in a notable BA loss.

\end{document}